\documentclass[10pt]{article}

\usepackage[footnotesep=0.15in,lmargin=0.75in,rmargin=0.75in,tmargin=1in,bmargin=1.7in]{geometry}
\usepackage{fancyhdr}
\usepackage{newtxtext}
\usepackage{titlesec}
\titleformat{\section}
  {\normalfont\fontsize{12}{15}\bfseries}{\thesection.}{1em}{}
\titleformat{\subsection}
  {\normalfont\fontsize{10.5}{12.6}\bfseries\slshape}{\thesubsection.}{1em}{}
\usepackage{booktabs}
\usepackage{tabu}
\usepackage{tabularx}

\usepackage[font=small,labelsep=period,labelfont=bf]{caption}
\usepackage[authoryear,sort]{natbib}
\newcommand{\minibio}[3]{
\includegraphics[width=0.16\textwidth]{#1}
\\
\textbf{#2} #3
\\
\hfill
}
\usepackage{eurosym}
\usepackage[caption=false]{subfig}
\usepackage{chngcntr}
\usepackage{rotating}
\usepackage{placeins}
\usepackage{bm}
\usepackage{float}
\usepackage{varioref}
\usepackage{makecell}
\usepackage[Symbolsmallscale]{upgreek}
\usepackage{multirow}
\usepackage{siunitx}
\usepackage{mathtools}
\usepackage[utf8]{inputenc}
\usepackage[T1]{fontenc}
\usepackage{hyperref}
\usepackage{url}
\usepackage{booktabs}
\usepackage{amsfonts}
\usepackage{nicefrac}
\usepackage{microtype}
\usepackage{amsmath}
\usepackage{amssymb}
\usepackage{graphicx}
\usepackage{array}
\usepackage{textcomp}
\usepackage{float}
\usepackage{enumitem}
\usepackage{comment}
\usepackage{soul}
\usepackage{color}
\usepackage{xspace}
\usepackage{cleveref}
\usepackage{empheq}
\usepackage[dvipsnames]{xcolor}

\newcommand{\tpp}{T_i \left(\dstr, \SD, \dperc, \capacity \right)}
\crefname{figure}{Fig.}{Fig.}
\Crefname{figure}{Fig.}{Fig.}

\newcommand{\vtts}{\nu}

\DeclareSIUnit{\year}{\text{year}}
\DeclareSIUnit{\million}{\text{million}}
\DeclareSIUnit{\EUR}{\text{\euro}}

\newcommand{\travtime}{c_{st}^{(1)}}
\newcommand{\waittime}{c_{rk}^{(2)}}
\newcommand{\odset}{Q}
\newcommand{\arcset}{E}
\newcommand{\nodeset}{V}
\newcommand{\shortminus}{{\scalebox{1.5}[1.5]{-}}}
\newcommand{\sigobs}{\SD_g}
\newcommand{\mape}{\text{MAPE}\left(\dstr, \SD\right)}
\newcommand{\rmsne}{\text{RMSNE}\left(\dstr, \SD\right)}
\newcommand{\triptime}{F}
\newcommand{\tloss}{T_{\text{loss}}}
\newcommand{\tgain}{T_{\text{gain}}}
\newcommand{\absgain}{\tgain\left(\dstr, \SD, \capacity, \dperc\right)}
\newcommand{\relgain}{\tgain^{\text{rel}}\left(\dstr, \SD, \capacity, \dperc\right)}
\newcommand{\optevalGT}{\triptime_{i}\left(\dstr, 0, \capacity, \dperc\right)}
\newcommand{\optevalstatic}{\triptime_{i}\left(\dstr, \SD, \capacity, 0\%\right)}
\newcommand{\absloss}{\tloss\left(\dstr, \SD, \capacity, \dperc\right)}
\newcommand{\relloss}{\tloss^{\text{rel}}\left(\dstr, \SD, \capacity, \dperc\right)}
\newcommand{\opteval}{F_{i}\left(\dstr, \SD, \capacity, \dperc\right)}
\newcommand{\tabapxref}[1]{Table \ref{#1}}
\newcommand{\dstr}{\mathcal{D}}
\newcommand{\normal}{\mathcal{N}}
\newcommand{\uni}{\mathcal{U}}
\newcommand{\expon}{\mathcal{E}}
\newcommand{\negexp}{-\mathcal{E}}
\newcommand{\weib}{\mathcal{W}}
\newcommand{\negweib}{-\mathcal{W}}
\newcommand{\capacity}{\gamma}
\newcommand{\fleetsz}{\pi}

\newcommand{\prtobs}[1]{\rho_{#1}^{\dstr, \SD}}
\newcommand{\SD}{\sigma}
\newcommand{\noisesample}[1]{\delta_{#1}^{\dstr, \SD}}
\newcommand{\equalnorm}[1]{$#1$}
\newcommand{\abovenorm}[1]{\emph{#1}}
\newcommand{\belownorm}[1]{$\bm{#1}$}
\floatstyle{ruled}

\newcolumntype{L}[1]{>{\raggedright\let\newline\\\arraybackslash\hspace{0pt}}m{#1}}
\newcolumntype{C}[1]{>{\centering\let\newline\\\arraybackslash\hspace{0pt}}m{#1}}
\newcolumntype{R}[1]{>{\raggedleft\let\newline\\\arraybackslash\hspace{0pt}}m{#1}}
\newcommand{\q}{\quad}
\newcommand{\qq}{\qquad}

\newcommand{\dperc}{\alpha}

\newcommand\numberthis{\addtocounter{equation}{1}\tag{\theequation}}
\newcommand{\trans}[1]{^{\mbox{\scriptsize T}}}

\DeclarePairedDelimiter\abs{\lvert}{\rvert}
\newcolumntype{P}[1]{>{\centering\arraybackslash}p{#1}}

\floatstyle{ruled}
\newfloat{Formulation}{tb}{myprog}

\newcommand{\fleet}{k = 1..\mfleetsz}
\newcommand{\od}{(o, d) \in \odset}
\newcommand{\arcs}{(s, t) \in \arcset}
\newcommand{\arcsrev}{(t, s) \in \arcset}
\newcommand{\routes}{r \in R}
\newcommand{\nodes}{s \in \nodeset_r}
\newcommand{\mfleetsz}{\overline{\fleetsz}}
\newcommand{\sm}[2]{\smashoperator{\sum_{#1}} #2}

\newcommand{\Flow}{x_{rst}^{od}}
\newcommand{\FlowReversed}{x_{rts}^{od}}
\newcommand{\FlowBoard}{b_{rks}^{od}}
\newcommand{\FlowAlight}{a_{rs}^{od}}
\newcommand{\RouteFlow}{y_{rk}}
\newcommand{\TravelDemand}{\prtobs{iod}}

\newcommand{\FleetSize}{\fleetsz}

\begin{document}

\begin{center}
% 18pt font for the title:
\fontsize{18pt}{21.6pt}\selectfont
\textbf{On the quality requirements of demand prediction for dynamic public transport}
\end{center}

% 10.5pt font for the entire document:
\fontsize{10.5pt}{12.6pt}\selectfont

{\ }

\begin{center}
Inon Peled$^{\text{*,a}}$, Kelvin Lee$^\text{b}$, Yu Jiang$^\text{**,a}$, Justin Dauwels$^\text{c}$, Francisco C. Pereira$^\text{a}$
\\
{\ \\}
$^\text{a}$ \emph{Danmarks Tekniske Universitet (DTU), Technology, Management and Economics Dept., Kgs. Lyngby, Denmark, 2800}
\\
$^\text{b}$ \emph{Nanyang Technological University (NTU), Graduate College, 50 Nanyang Avenue, Singapore, 637553}
\\
$^\text{c}$ \emph{Delft University of Technology (TU Delft), Microelectronics Dept., the Netherlands, 2600}
\end{center}

{\ }

\begin{center}
* Corresponding author\ \ \ \ \ ** Co-corresponding author
\\
\emph{E-mail addresses: $\{\text{inonpe,yujiang}\}$@dtu.dk}
\end{center}

\section*{Abstract}
As Public Transport (PT) becomes more dynamic and demand-responsive, it increasingly depends on predictions of transport demand.
But how accurate need such predictions be for effective PT operation?
We address this question through an experimental case study of PT trips in Metropolitan Copenhagen, Denmark, which we conduct independently of any specific prediction models.
First, we simulate errors in demand prediction through unbiased noise distributions that vary considerably in shape.
Using the noisy predictions, we then simulate and optimize demand-responsive PT fleets via a linear programming formulation and measure their performance.
Our results suggest that the optimized performance is mainly affected by the skew of the noise distribution and the presence of infrequently large prediction errors.
In particular, the optimized performance can improve under non-Gaussian vs. Gaussian noise.
We also find that dynamic routing could reduce trip time by at least $23\%$ vs. static routing.
This reduction is estimated at $\SI{809000}{\EUR\per\year}$ in terms of Value of Travel Time Savings for the case study.

\section*{Keywords}
Dynamic public transport; Demand forecasting; Non-Gaussian noise; Predictive optimization.

\section{Introduction} \label{sec:introduction}

Public Transport (PT) has traditionally used static itineraries that remain unchanged for months \citep{ceder2016public}.
However, as autonomous mobility advances and the vision of Smart Cities takes shape, the day approaches when PT becomes dynamic, so that some itineraries are adapted to real-time transport demand (i.e., are demand-responsive) \citep{bosch2018cost, horavzvdovsky2018data}.
Meanwhile also, predictive models of transport demand are increasingly used for both long-term and short-term traffic management \citep{hashemi2015real}.
Future Public Transport should thus naturally employ predictive models for timely adaptation of service per expected transport demand.

The effective operation of demand-responsive PT thus requires that transport demand be accurately estimated ahead of time.
For example, more accurate demand predictions can yield better utilization of PT resources, e.g., so that fewer vehicles are used while travel times are also cut shorter.
Conversely, errors in predicted demand might lead to sub-optimal routing, thus resulting in the waste of energy, longer waiting times and profit loss.

The main goal of this work is to study the impact of demand prediction accuracy on subsequent performance of demand-responsive PT.
To this end, we conduct a hypothetical case study of a demand-responsive PT ``pilot'' experiment in Metropolitan Copenhagen, Denmark.
We do so by first simulating prediction errors through various stochastic perturbations of travel demand, as observed through real-world PT trips.
Using the noisy predictions, we then simulate and optimize demand-responsive fleets via a linear programming formulation, for various bus capacities and portions of dynamically routed buses.
Finally, we analyze the results and draw conclusions on the impact of prediction accuracy on optimization quality.

Our case study is thus a form of Sensitivity Analysis, as we review in the next Section.
However, contrary to most other Sensitivity Analysis studies on demand-responsive PT optimization, we account for both sides of demand prediction and dynamic routing and do so separately.
Also in contrast to previous works, we consider predictive noise independently of any specific demand estimation models, to study the impact of prediction errors once any such model has already been fitted.
We use different types of independent noise, both Gaussian and non-Gaussian, with a wide variety of statistical parameter settings.
This allows us to investigate various correlation structures in the origin-destination pairs, e.g., as associated with non-recurrent traffic disruptions.

{
The motivation for this work is thus to offer several novelties over previous research into emergent, dynamically routed PT services.
On the methodological side, we take a high-level view of the impact of predictive accuracy on such PT, rather than use specific prediction models as in other works.
This allows us to study such impacts without binding to any particular family of models and its specific properties as in other works.
For example, models that rely on lagged information (autoregressive models, Recurrent Neural Networks, etc.) generally share some similarities in the effect of such parameters on predictive accuracy, and so our goal is to study the error distributions themselves, rather than particular parameters, learning methods, hyper-parameter tuning, etc.
In turn, this also facilitates an examination of the relationship between optimal service quality and deviations from Gaussian predictive noise, as commonly assumed.
On the practical side, this methodology allows us to explicitly quantify gains and losses in optimizing such services under a wide variety of possible noise distributions, without committing to a particular source of noise.
That is, while predictive noise can arise from multiple sources -- e.g., modeling choices, non-recurrent traffic disruptions and changes in mode preferences -- we quantify its practical impact on the optimized service regardless of its sources.
}

The remainder of this work is organized as follows.
Section \ref{sec:literature} reviews gaps in current studies on demand-responsive PT optimization and uncertainty analysis.
Section \ref{sec:experiments} details our experiments with simulated noise distributions and prediction-based optimization.
Section \ref{sec:results} then provides the experimental results and their analysis.
Finally, Section \ref{sec:discuss} recaps the work by discussing the goal, methodology and main results, and Section \ref{sec:conclude} concludes with a list of key findings and future steps.

\section{Literature review} \label{sec:literature}

\subsection{Public transport optimization}

Works on PT planning and design rely traditionally on point estimates of future travel demand, obtained through manual data collection (e.g., via transport surveys).
These estimates are thus realizations from a latent (i.e., unknown) transport demand distribution, and so are subject to uncertainty and errors.
While such errors are widely acknowledged, their subsequent effects on PT performance are less quantified and discussed.

Modern works on PT optimization take advantage of advancements in big data and machine learning for demand estimation \citep{KRISHNAKUMARI202038,TOOLE2015162} and real-time fleet management \citep{9091900}.\footnote{Ref. \citep{Iliopoulou2019} for a comprehensive review of big data applications in PT.}
As these advancements gain traction in the PT optimization field \citep{Wang2019, Iliopoulou2019}, it becomes increasingly important to study the effects of demand prediction accuracy on optimization quality.
Further, these effects should also be evaluated in the context of future PT, which will employ dynamically routed vehicles for better demand-responsiveness \citep{bosch2018cost}.

Motivated by these needs, we experiment with a case study of a PT fleet, whose itineraries are dynamically changed per predicted travel demand, in hourly intervals.
We optimize service times under constraints of passenger preferences and service quality via Mixed-Integer Linear Programming (MILP), as investigated in many PT optimization studies \citep{wang2010global,luathep2011global,tong2015transportation,SZETO2014235,AN2015737}.

The fleet in this work thus resembles Mobility-on-Demand services and similarly relies on predicted transport demand, albeit with less flexible routes and stop locations.
However, whereas many such services operate on the basis of pre-booked rides \citep{8206203,HYLAND2018278}, we do not assume any particular source of demand observations.
Accordingly, PT demands in this case study are observed only upon their realization, i.e., only when passengers board or alight.

Contrary to previous works on PT optimization under demand uncertainty \citep{ukkusuri2007robust}, we study a wide range of possible distribution properties (e.g., skew and kurtosis).
Some works further attempt to mitigate errors via robust optimization techniques \citep{AN2015737} and chance constraints \citep{wang2015sustainable}.
As error mitigation is not the focus of our case study, we consider this and other extensions for future work (Section \ref{sec:conclude}).

\subsection{Sensitivity analysis for demand-responsive public transport}

By quantifying the impact of changes in demand prediction accuracy on subsequent PT performance, our case study is a form of Sensitivity Analysis (SA) \citep{saltelli2004sensitivity}.
Nevertheless, this work differs in several respects from other SA studies in the field of demand-responsive PT optimization, as follows.

First, most SA studies on demand-responsive PT either treat demand as given or change it per predefined levels.
For example, \citet{huang2020flexible} conduct an SA study for operating a demand-responsive transit service under deterministic changes in total demand, while \citet{WINTER2018151} do so for an automated demand-responsive transport system.
Manasra and Toledo (\citeyear{manasra2019optimization}) incorporate demand predictions within an optimization formulation and measure its sensitivity to predefined changes in demand.
They note that for practical usefulness, the robustness of their formulation needs to be further studied under non-recurrent disruptions and demand surges.

Second, SA studies on demand-responsive transit often do not explicitly deal with stochastic noise in demand estimation.
For example, \citet{nickkar2020sensitivity} measure the effects of varying demand on performance of autonomous fleets, but do not deal with errors in demand estimation.
Similarly, \citet{lee2006transit} optimizes a transit network under various pre-defined scenarios, without checking possible deviations from optimality due to demand estimation errors.
\citet{ibeas2014bus} optimize bus capacity and headway per elastic demand, as collected from historical records, and without simulating stochastic noise.

Contrary to the above, we conduct SA under stochastic errors in demand estimation for demand-responsive PT.
Moreover, the above works focus on developing applicable formulations for PT optimization, which motivates their choice of conducting SA under predefined levels of changes.
Conversely, we do not aim to propose a directly applicable PT optimization method, but rather use the simulated PT system to study predictive noise effects from a more general perspective, as we next explain.

\subsection{Uncertainty in modeling and the normality assumption}

Predictive noise follows from the existence of uncertainty in modeling \citep{beaudrie2016using} and yields residuals, i.e., differences between modeled predictions and actually observed values\footnote{Definitions of residuals differ by context and specificity \citep{gourieroux1987generalised}; we use a general definition that befits the context in this work.}.
Consequently, when fitting predictive models, a normality assumption is commonly employed, whereby residuals are expected to be identically and Normally distributed \citep{seber2012linear}.
In some modeling contexts, however, residuals can be non-Gaussian \citep{mak2000heteroscedastic}.
It may also be impractical to fit a model with Gaussian residuals, depending on data size and quality \citep{jackson2018meta}.

The probability density of residuals can thus vary considerably in standard deviation (dispersion around the mean), skew (asymmetry around the mean) and kurtosis (weight of tails).
Multiple methods have been devised for \emph{detecting} deviations from normality, including plots, comparison of moments, and statistical tests \citep{thode2002testing}.
However, there are far fewer works on the \emph{impact} of such deviations \citep{ivkovic2020coverage}, most of which concentrate on errors in model parameter estimates.

Nelson and Granger (\citeyear{nelson1979experience}) discovered that for economic time series, Autoregressive Moving Average (ARMA) models often yield residuals with markedly non-Gaussian skew and kurtosis.
\citet{davies1980autoregressive} expanded on this and concluded that the use of non-Gaussian residuals actually allows for a larger selection of models that better represent time series.
In the context of chemical analysis, Wolters and Kateman (\citeyear{wolters1989performance}) used Monte Carlo simulations to quantify errors in Least Squares parameters under small deviations from normality.

More recently, He and Raghunathan (\citeyear{he2009imputation}) use simulated data to examine sequential regression imputation methods under shifted and scaled non-Gaussian distributions: Uniform, Lognormal and $t$-distribution.
They find that mean performance remains quite robust, despite noticeable instability in regression coefficients.
To predict debt and bankruptcy of Serbian companies, \citet{ivkovic2020coverage} simulate Exponential and Weibull distributed residuals, study the resulting errors in LR parameter estimates, and devise transformations to reduce these errors.
\citet{pernot2020impact} study non-Gaussian errors in Quantum Machine Learning models and find that mean measures of prediction error depend significantly on the shape of the error distribution.
They also note the need for more research into the impact of error distributions on model reliability in general.

Similarly to the above studies, this work uses simulated perturbations of real data to study the impact of prediction errors under different noise distributions.
However, none of the previous studies apply directly to the transport domain, and most of them deal with deviations from normality in linear modeling.
In contrast, we neither presume any particular modeling form nor try to mitigate uncertainty, but rather examine its effect on demand-responsive PT.
Per common modeling practices, we assume only that the predictive model has low bias and is evaluated through mean error measures, as detailed in the following Sections.

\section{Experiments} \label{sec:experiments}

\subsection{Data}

\begin{figure}[tb]
\centering
\includegraphics[width=0.5\linewidth]{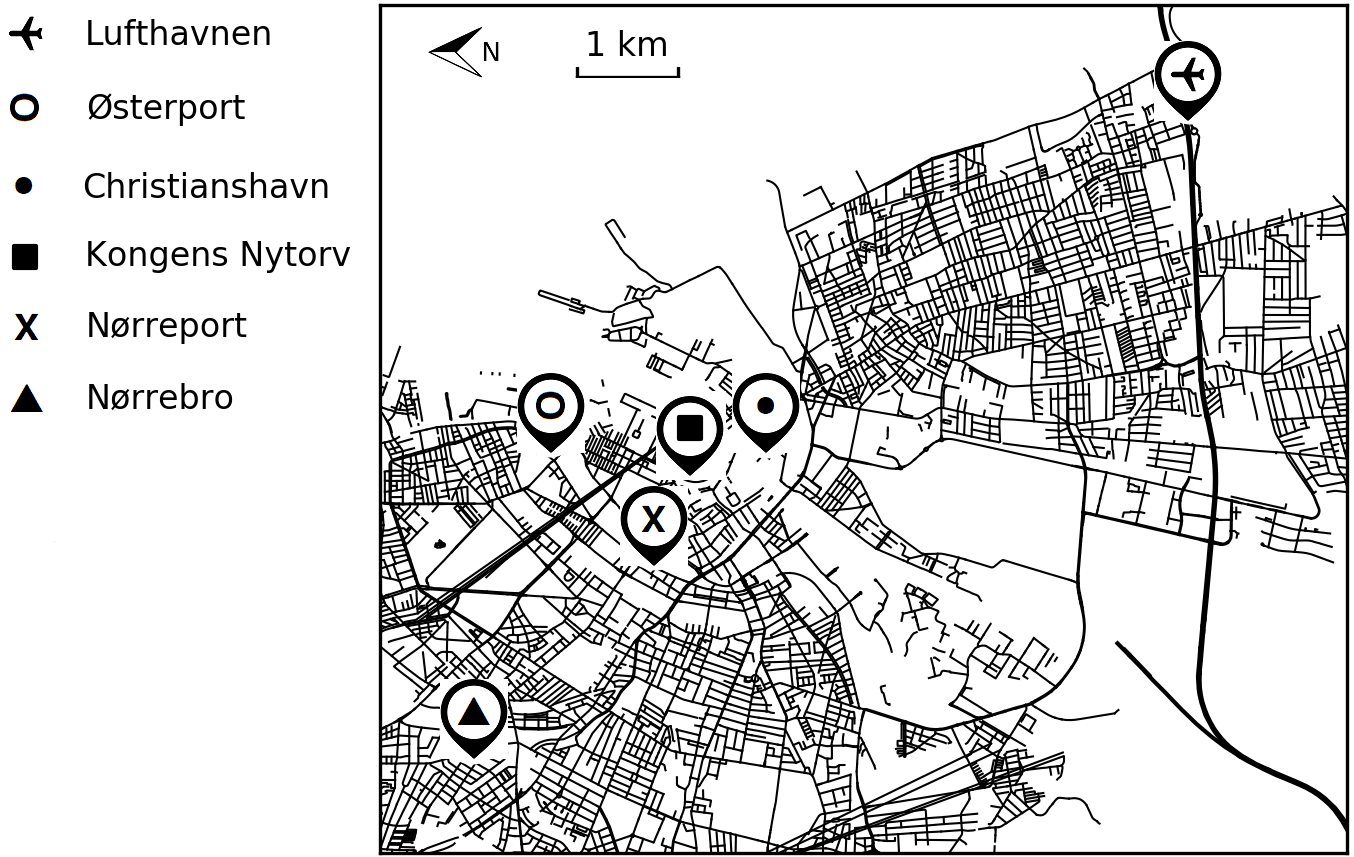}
\caption{The $6$ most active PT stations in Metropolitan Copenhagen, Denmark.}
\label{fig:stations}
\end{figure}

The data for the case study consists of PT trips from 1-Jan-2017 to 21-Dec-2018 in Metropolitan Copenhagen, Denmark.
These trips are conducted with electronic travel cards (``Rejsekort'') and so account for approximately $1 / 3$ of all bus and train trips.
As dynamically routed Public Transport (PT) is still emerging, we consider a hypothetical ``pilot'' experiment, where dynamically routed buses serve the $6$ most active PT stations, depicted in \Cref{fig:stations}.
For these $6^2 - 6 = 30$ Origin-Destination (OD) pairs, the data contains $\SI{2.15}{\million}$ trips, based on which we simulate various closed PT systems with urban-scale travel distances.

We aggregate the data by counting hourly trip starts for each OD pair.
Then, we draw at random $N = 100$ hours from 2018-Dec-1 00:00, $\dots$, 2018-Dec-21 23:00.
The Ground Truth observations are thus the hourly trip counts for each OD pair in these $N$ hours.
Based on the Ground Truth observations, we next simulate noisy predictions of PT demand.

\subsection{Generation of noisy demand predictions} \label{sec:noisegen}

Similarly to the bulk of existing literature on non-Gaussian residuals (Section \ref{sec:literature}), we assume that the noise is generated from continuous distributions.
As such, negative PT demand values are possible and could indicate preference to use other transport modes (e.g., bike, car, walking, etc.) over PT.
Further, we use independent noise distributions, to deal with the challenging case of unpredictable changes in correlation structure.
For instance, given two OD's that are typically positively correlated, a non-recurrent disruption (e.g., a road block) might cause a surge in the usage of one of them along with a steep decline in the other.
The noise distributions we experiment with are all homoscedastic with standard deviation (SD) $\SD$, for $\SD = 0.5, 1.0, 2.0, 3.0$ \footnote{
    We have also experimented with a finer grained range of $\SD$ and obtained consistent results, which we thus omit for brevity.
}, as follows.
\begin{enumerate}
\itemsep1pt
\item Gaussian, i.e., Normal: $\normal\left(0, \SD^2\right)$.
\item Uniform: $\uni\left[0, \sqrt{12}\SD\right]$.
\item Exponential ($\expon$) with scale $\SD$.
\item Negated Exponential, namely, $\negexp$.
\item Weibull ($\weib$) with scale $1$ and shape that corresponds to SD $\SD$; reduces to $\expon$ for  $\SD = 1$.
\item Negated Weibull, namely, $\negweib$.
\end{enumerate}

\begin{figure}[tb]
    \centering
    \includegraphics[width=\linewidth]{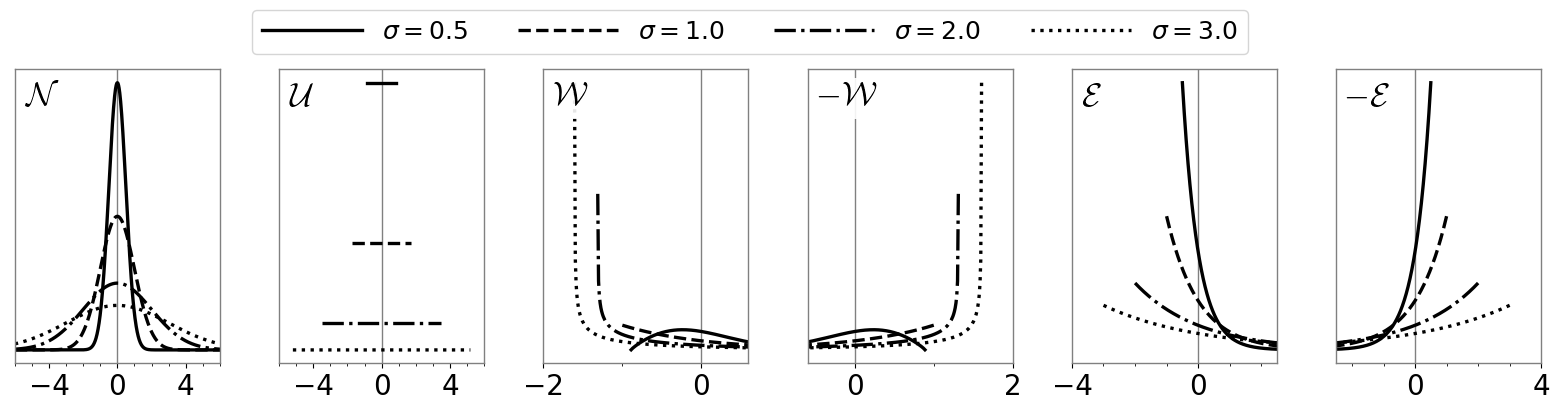}
    \caption{Probability Density Functions (PDFs) of the noise distributions with zero mean.
    Note the different horizontal scale and support of each distribution.
    Vertical scales differ too and are omitted, because the density value is inconsequential in this work.}
    \label{fig:pdfs}
\end{figure}

We shift back each distribution by its expected value to obtain zero mean, as illustrated in \Cref{fig:pdfs}.
\Cref{tab:skewkurt} compares the noise distributions with $\mathcal{N}\left(0, 1\right)$, the standard Gaussian, in terms of their 3rd and 4th standardized moments, namely, skew and kurtosis.
This Table shows that our experiments cover a range of distribution properties: platykurtic, leptopkurtic and mesokurtic -- i.e., having kurtosis below, above, or equal to $\mathcal{N}\left(0, 1\right)$ -- as well as positive, negative and zero skew.
The distributions also vary in PDF support, which is either finite ($\uni$), semi-infinite ($\expon$, $\negexp$, $\weib$, $\negweib$) or infinite ($\normal$).

For each OD pair $(o, d)$, shifted noise distribution $\dstr$ and standard deviation $\SD$, we independently draw $N$ samples, $\noisesample{1od}, \dots, \noisesample{Nod} \sim \dstr$.
Then for all $i=1 \dots N$, we let $g_{iod}$ denote the corresponding Ground Truth observation, and generate noisy predictions as follows:
\begin{equation}
\prtobs{iod} \coloneqq g_{iod} + \sigobs \cdot \noisesample{iod}
\,,
\label{eq:noise}
\end{equation}
where $\sigobs$ is the sample SD of all ground truth observations.
The stochastic noise is thus measured in units of the SD of the observations themselves.
Furthermore, every $\prtobs{iod}$ is unbiased, as all noise distributions have zero mean.

\begin{table}[tb]
    \renewcommand{\arraystretch}{0.9}
    \caption{Skew and kurtosis of noise distributions, marked in comparison to $\normal(0, 1)$ as: \textbf{below}, \emph{above}, equal.}
    \label{tab:skewkurt}
    \centering
    \begin{tabular}{|c|c|c|c|c|c|c|c|}
    \cline{1-8}
	$\SD$ & Property & \thead{$\normal$} & 	\thead{$\uni$} & \thead{$\expon$} & 	\thead{$\negexp$} & \thead{$\weib$} & 	\thead{$\negweib$} \\
	\cline{1-8}
\multirow{2}{*}{$0.5$} & Skew & \equalnorm{0.00} & \abovenorm{1.30} & \abovenorm{0.75} & \belownorm{\shortminus0.75} & \abovenorm{1.46} & \belownorm{\shortminus1.46} \\
 & Kurtosis & \belownorm{0.19} & \belownorm{1.80} & \belownorm{1.50} & \belownorm{1.50} & \belownorm{2.35} & \belownorm{2.35} \\
 \cline{1-8}
\multirow{2}{*}{$1.0$} & Skew & \equalnorm{0.00} & \abovenorm{10.39} & \abovenorm{6.00} & \belownorm{\shortminus6.00} & \abovenorm{6.00} & \belownorm{\shortminus6.00} \\
 & Kurtosis & \equalnorm{3.00} & \abovenorm{28.80} & \abovenorm{24.00} & \abovenorm{24.00} & \abovenorm{24.00} & \abovenorm{24.00} \\
 \cline{1-8}
\multirow{2}{*}{$2.0$} & Skew & \equalnorm{0.00} & \abovenorm{83.14} & \abovenorm{48.00} & \belownorm{\shortminus48.00} & \abovenorm{47.75} & \belownorm{\shortminus47.75} \\
 & Kurtosis & \abovenorm{48.00} & \abovenorm{460.80} & \abovenorm{384.00} & \abovenorm{384.00} & \abovenorm{624.50} & \abovenorm{624.50} \\
 \cline{1-8}
\multirow{2}{*}{$3.0$} & Skew & \equalnorm{0.00} & \abovenorm{280.59} & \abovenorm{162.00} & \belownorm{\shortminus162.00} & \abovenorm{183.31} & \belownorm{\shortminus183.31} \\
 & Kurtosis & \abovenorm{243.00} & \abovenorm{2,332.80} & \abovenorm{1,944.00} & \abovenorm{1,944.00} & \abovenorm{4,975.26} & \abovenorm{4,975.26} \\
     \cline{1-8}
    \end{tabular}
\end{table}

\subsection{Fleet optimization} \label{sec:opt}

Having generated noisy predictions of PT demand, we now simulate the demand-responsive PT services with varying no. of buses, bus capacity, and percentage of dynamically routed buses.
For capacity, our reference value is $110 / 3 \approx 33$, as the average bus capacity in Metropolitan Copenhagen is approx. $110$, and Rejsekort accounts for approx. $1 / 3$ of all PT trips.
We vary capacity around this value as $\capacity = 10, 20, 30, 40$; the results later show that this provides sufficient insight into the effect of capacity variability.
For percentage of dynamically routed buses, we use $\dperc = 0\%, 10\%, \dots, 40\%$.
In particular, $\dperc = 0\%$ corresponds to a completely statically routed fleet, whose performance is thus independent of predictions and noise.
{
Table {\ref{tab:triptime}} later shows that the mean and standard deviation of trip time per passenger remain stable as $\dperc$ increases from $30\%$ to $40\%$, hence we do not increase $\dperc$ further.
For this case study then, conversion to dynamic routing beyond $\dperc = 40\%$ does not yield significantly better routes.
}

Finally, we solve a mixed-integer linear programming (MILP) formulation to obtain optimal routes {and fleet size}.
This formulation receives as input a directed graph, where nodes correspond to PT stations, and an arc between two nodes exists if a bus can travel between the two stations.
Each edge is weighted by the corresponding in-vehicle travel time.
The input also includes a set of bi-directional routes $R$, where for every combination of $2$ or more nodes, $R$ contains the shortest acyclic path connecting these nodes.
{
Given a total of $\mfleetsz$ buses, the objective is to minimize both the number of buses actually deployed and passengers' total trip time (i.e., waiting time and in-vehicle travel time) in a given time horizon, by assigning routes to $\FleetSize \leq \mfleetsz$ of the buses.
In particular, we constrain $\FleetSize$ to fulfill two separate requirements: 1) $\FleetSize$ buses can satisfy the predicted demand in the given time horizon, 2) $\FleetSize$ buses can satisfy demand even in the $100$ historically worst-case scenarios -- i.e., the $100$ busiest hours, by number of trips, before 1-Dec-2018 -- if they are routed to serve all stops.
In this manner, the optimized fleet is likely to be feasible for the actual, ground truth demand in the given time horizon.
}

We assume that passengers choose the shortest route from origin to destination, and waiting times and in-vehicle travel times are prespecified.
The optimizer first truncates any negative predictions to zero, as negative demand implies no passengers.
Note that the noise PDFs themselves are not truncated, but rather the predictions derived from them.

\begin{table}[tb]
    \renewcommand{\arraystretch}{0.95}
    \centering
    \caption{Notation for \autoref{prog:opt}.}
    \label{tab:optnotat}
    \begin{tabular}{|ll|l|}
        \cline{1-3}
        \multirow{4}{*}{\textbf{Sets}}
        & $\nodeset$              & nodes\\
        & $\arcset$               & arcs\\
        & $\odset$				& OD pairs\\
		& $R$						& routes\\
		& $\nodeset_r$              & nodes in route $r$\\
		\cline{1-3}
		\multirow{5}{*}{\textbf{Indices}}
		& $s, t$     & nodes\\
        & $(o, d)$     & OD pair\\
        & $r$ & route \\
        & $\underline{r}$   & base route that serves all nodes \\
        & $k$ & no. buses allocated to a route \\
        \cline{1-3}
        \multirow{5}{*}[15pt]{\makecell{\ \\\textbf{Parameters}}}
        & $\dperc$         & portion of buses that can be dynamically routed \\
        & $\capacity$                  & capacity of each bus \\        
        & $\travtime$             & in-vehicle travel time from $s$ to $t$\\
        & $\waittime$             & average waiting time for route $r$, when $r$ is allocated $k$ buses\\
        & $M$                     & large number \\
        & $C$                     & operational cost of each bus\\
        & $\mfleetsz$             & maximum fleet size\\
        \cline{1-3}
        \multirow{4}{*}[15pt]{\makecell{\ \\ \ \\\hspace{-3pt}\textbf{Decision}\\\textbf{Variables}}}
        & $\fleetsz$              & fleet size\\
        & $\Flow$ & flow of passengers traveling from $s$ to $t$ in route $r$, as part of a trip from $o$ to $d$ \\
        & $\FlowBoard$            & flow of passengers who board route $r$ at $s$, as part of a trip from $o$ to $d$, when $r$ is allocated $k$ buses\\ 
        & $\FlowAlight$			& flow of passengers who alight from route $r$ at $s$, as part of a trip from $o$ to $d$ \\
        & $\RouteFlow$			& binary indicator of allocation of $k$ buses to route $r$\\
        \cline{1-3}
    \end{tabular}
\end{table}

\begin{Formulation}[H]
\caption{Fleet Optimization} \label{prog:opt}
\begin{flalign}
	& \text{Minimize} \notag \\
	& C \FleetSize + \q \sm{\substack{\od,\routes,\\\arcs}}{\travtime \Flow} \q + \q \sm{\substack{\od,\routes,\\\nodes,\fleet}}{\waittime \FlowBoard} \label{eq:obj}
		& & \hspace{0.5cm} \\
	& \text{subject to} \notag \\
	& \sm{\q \fleet,}{\FlowBoard} + \sm{t: \arcs}{\Flow} - \sm{t: \arcsrev}{\FlowReversed} - \FlowAlight = 0 \label{constr:flowconservation}
		& \forall \od, \q \routes, \q \nodes \\
    & \sm{\substack{\routes,\\\q\fleet}}{\FlowBoard} - \sm{\routes}{\FlowAlight} = \begin{cases}
 		\TravelDemand & \text{$s = o$}\\
 		-\TravelDemand & \text{$s = d$}\\
 		0 & \text{$s \notin \{o, d\}$}
 	\end{cases} \label{constr:demands}
		& \forall \od, \q \nodes \\
	& \sm{\substack{\q \qq \nodes, \od}}{\FlowBoard - M \RouteFlow \leq 0} \label{constr:definitiony}
		& \forall \routes, \q \fleet \\
	& \sm{\q \od}{\Flow} - \capacity k \RouteFlow \leq 0 \label{constr:capacity}
		& \forall \routes, \q \arcs \\
	& \sm{\q \fleet}{\RouteFlow} \leq 1 \label{constr:fleet2}
		& \forall \routes \\
	& \sm{\q \qq\routes, \fleet}{k \RouteFlow} - \FleetSize \leq 0 \label{constr:fleet}
		& \\
	& \sm{\q \fleet}{k y_{\underline{r}k}} - (1 - \alpha) \FleetSize \geq 0 \label{constr:base}
		& \forall \fleet \\
    & \fleetsz \leq \mfleetsz \label{constr:fleetsize}\\
	& b_{j\underline{r}\FleetSize s}^{od} + \sm{t: \arcs}{x_{j\underline{r}st}^{od}} - \sm{t: \arcsrev}{x_{j\underline{r}ts}^{od}} - a_{j\underline{r}s}^{od} = \begin{cases}
 		\rho_{jod} & \text{$s = o$}\\
 		-\rho_{jod} & \text{$s = d$}\\
 		0 & \text{$s \notin \{o, d\}$}
 	\end{cases} \label{constr:flowconservationbusiest}
		& \forall \od, \q \nodes, \q j = 1...100 \\
	& \sm{\q \od}{x_{j\underline{r}st}^{od}} - \capacity \FleetSize \leq 0 \label{constr:capacitybusiest}
		& \forall \arcs, \q j = 1...100\\
	& \RouteFlow \in \{0, 1\}, \q \Flow, \FlowBoard, \FlowAlight, \FleetSize \geq 0, \q \FleetSize \text{ is integral} \label{constr:domain}
		& \forall \routes, \q s, t \in V_{r}, \q \od, \q \fleet
\end{flalign}
\end{Formulation}

\FloatBarrier

For any hour $i=1 \dots N$, \autoref{prog:opt} defines the linear program per the notation in \Cref{tab:optnotat}.
The objective and constraints are similar to those commonly used for fleet optimization, as follows.
In objective \eqref{eq:obj}, {the first term is the operational cost of deploying a fleet of $\fleetsz$ buses.
We use a large cost coefficient $C$ per bus, to obtain the smallest fleet that still satisfies the demands.
The second term is the total passengers' in-vehicle travel time, and the third term is total passengers' waiting time.
{Constraint \eqref{constr:flowconservation}} imposes flow conservation, such that the flow into any node of any route (from any previous node in the route) either proceeds into the next node in the route or flows out of the route through $a_{rs}^{od}$.
This ensures that the itinerary of each passenger is complete from their origin to their destination.
Constraint {\eqref{constr:demands}} sets the flow per boarding and alighting at a stop $s$.
This ensures that the total flow for OD pair $(o, d)$ into node $o$ across all routes sums to the demand, and the total flow for OD pair $(o, d)$ out of node $d$ across all routes sum to the same demand.
Constraint {\eqref{constr:definitiony}} ensures that when a passenger boards any route, the route and the corresponding no. of buses are selected, i.e., $y_{rk} = 1$.
For example, if the flow $b_{rks}^{od}$ is non-zero, then $y_{rk}$ must be equal to one, indicating that exactly $k$ buses are allocated to route $r$.
Constraint {\eqref{constr:capacity}} defines the capacity along all segments of all routes.
Constraint {\eqref{constr:fleet2}} ensures that only one $y_{rk}$ is set to $1$ for each route, such that the number of buses allocated to each route is deterministic.
}
Constraint \eqref{constr:fleet} then guarantees {that the total number of buses allocated to all routes does not exceed the deployed fleet size $\FleetSize$}, while constraint \eqref{constr:base} dictates that at least $(1 - \alpha)\pi$ buses serve route $\underline{r}$, which is route that serves all stops.
{
Constraint {\eqref{constr:fleetsize}} imposes the maximum fleet size.
}
Constraints \eqref{constr:flowconservationbusiest} and \eqref{constr:capacitybusiest} {ensure that the fleet of $\pi$ buses, when routed to serve all stops, can satisfy the demand in each of the $100$ busiest hours before 1-Dec-2018 (denoted by subscript $j$).}
Finally, constraint \eqref{constr:domain} defines the possible set of values for each decision variable.
We use $M = \num{100000}$, $C = \num{100000}$ and $\mfleetsz = 40$ in this case study.

\section{Results} \label{sec:results}
The optimized fleet sizes $\fleetsz = 38, 19, 13, 10$ are consistent across all generated noisy demands for $\capacity = 10, 20, 30, 40$, respectively.
\autoref{tab:ds} further specifies the no. dynamically and statically routed buses for each $\capacity$ and $\dperc$.
This percentage is seen to correspond to the PDFs in \autoref{fig:pdfs}; e.g., as $\SD$ increases, $\negweib$ yields the fewest negative predictions as its PDF shifts towards the positives.

\begin{table}[bt]
    \caption{No. dynamically (D) and statically routed (S) buses.}
    \label{tab:ds}
    \centering
    \begin{tabular}{|c|cc|cc|cc|cc|cc|}
    \cline{1-11}
    & \multicolumn{2}{c|}{$\dperc=0\%$} & \multicolumn{2}{c|}{$\dperc=10\%$} & \multicolumn{2}{c|}{$\dperc=20\%$} & \multicolumn{2}{c|}{$\dperc=30\%$} & \multicolumn{2}{c|}{$\dperc=40\%$} \\
    $\capacity$ & D & S & D & S & D & S & D & S & D & S \\
    \cline{1-11}
10 & 0 & 38 & 4 & 34 & 8 & 30 & 12 & 26 & 16 & 22 \\
20 & 0 & 19 & 2 & 17 & 4 & 15 & 6 & 13 & 8 & 11 \\
30 & 0 & 13 & 2 & 11 & 3 & 10 & 4 & 9 & 6 & 7 \\
40 & 0 & 10 & 1 & 9 & 2 & 8 & 3 & 7 & 4 & 6 \\
\cline{1-11}
    \end{tabular}
\end{table}

We evaluate prediction quality through two commonly used, unitless measures of mean error: Mean Absolute Percentage Error (MAPE) and Rooted Mean Squared Normalized Error (RMSNE).
These measures are defined as follows:
\begin{align*}
    \mape 
    &\coloneqq \frac{1}{N} \sum_{i=1}^N \frac{1}{\abs{\odset}} \sum_{(o, d) \in \odset} \frac{\abs{\prtobs{iod} - g_{iod}}}{\bar{g}}\\
    &= \frac{\sigobs}{\bar{g} \abs{\odset} N} \sum_{i=1}^N \sum_{(o, d) \in Q}
    \abs{\noisesample{iod}}
    \,, \numberthis \label{eq:mape}
    \\
    \rmsne 
    &\coloneqq \sqrt{\frac{1}{N} \sum_{i=1}^N \frac{1}{\abs{\odset}} \sum_{(o, d) \in \odset} \left(\frac{\prtobs{iod} - g_{iod}}{\bar{g}}\right)^2}
    \\
    &= \frac{\sigobs}{\bar{g} \sqrt{\abs{\odset} N}} \sqrt{\sum_{i=1}^N \sum_{(o, d) \in \odset} {\left(\noisesample{iod}\right)^2}}
    \,, \numberthis \label{eq:rmsne}
\end{align*}
where $\bar{g}$ is the mean of all ground truth observations.
Each right-hand side follows from \eqref{eq:noise} and shows the dependency on sampled noise (where the fractional coefficient is noise-independent).

\begin{figure}[tb]
\centering
\includegraphics[width=0.65\linewidth]{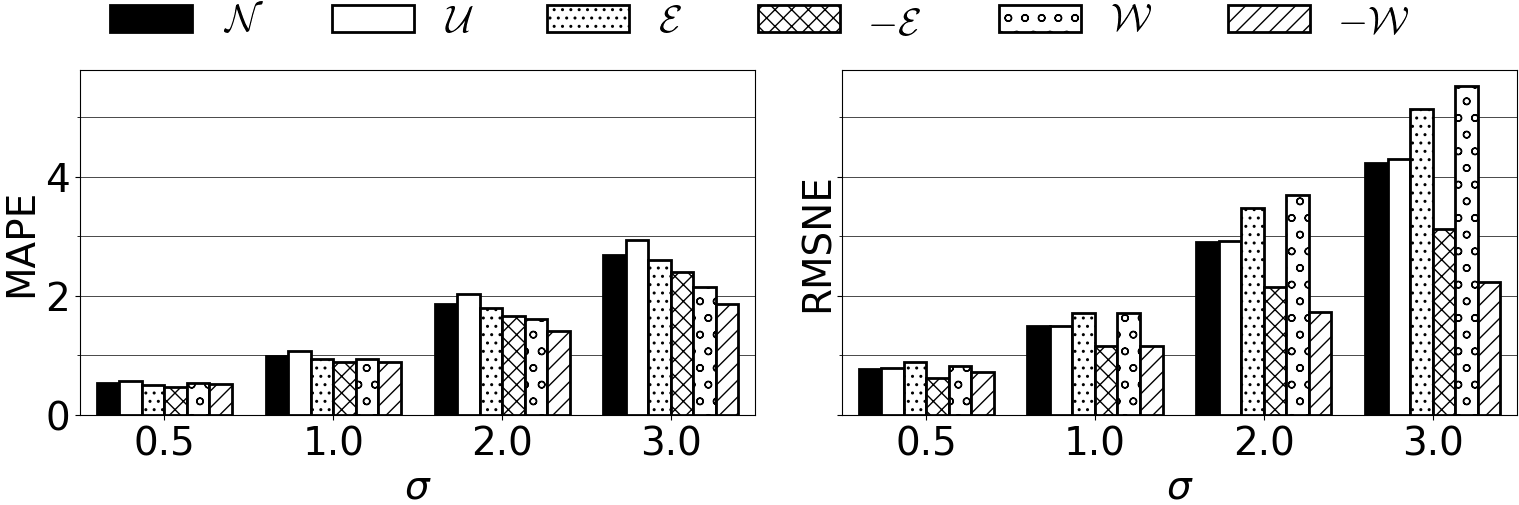}
\caption{MAPE and RMSNE after the optimizer truncates negative predictions. Lower is better.}
\label{fig:predaftr}
\end{figure}

\Cref{fig:predaftr} illustrates the MAPE and RMSNE after the optimizer truncates negative predictions.
As expected, both MAPE and RMSNE increase as $\SD$ increases.
When ranking $\dstr$ from best to worst, we obtain
$\negweib \leq \weib \leq \negexp \leq \expon \leq \normal \leq \uni$ for MAPE vs. $\negweib \leq \negexp \leq \normal \leq \uni \leq \expon \leq \weib$ for RMSNE.
Let us next analyze how closely the optimization performance follows any of these rankings.

We evaluate optimization performance through several measures, all of which are based on the objective value (i.e., total trip time).
For generalizability, our results are mostly given in relative terms.
However, to give a basic sense of scale for this case study, we begin with an absolute measure of trip time ($\si{\min}$) per passenger.
For this, let
\begin{align}
    \tpp \coloneqq \frac{\opteval}{P_i}
    \,,
    \label{eq:triptime}
\end{align}
where for all $i=1 \dots N$, $P_i$ is total no. passengers observed in the $i$'th hour, and $\opteval$ is their total trip time using the optimized fleet.

\Cref{tab:triptime} summarizes $\tpp$ through its mean and SD over all $i, \dstr, \SD$.
The mean and SD are both seen to decrease as $\capacity$ decreases and/or $\dperc$ increases.
Hence as expected, trip times improve if more buses with lower capacity are used and/or more of the buses are dynamically routed.
We also note that the mean and SD do not vary much when further separating by $\dstr$ and $\SD$, for any fixed $\dperc \,, \capacity$.
This is a possible consequence of using the same marginal noise distribution for all OD pairs, so that their predicted demands shift similarly, thereby balancing each other out during fleet optimization.

\begin{table}
\centering
\caption{Mean ($\pm \text{SD}$) of $\tpp$, in minutes. Lower is better.}
\label{tab:triptime}
\begin{tabular}{|c|c|c|c|c|}
\cline{1-5}
\thead{$\dperc$} & \thead{$\capacity = 10$} & \thead{$\capacity = 20$} & \thead{$\capacity = 30$} & \thead{$\capacity = 40$} \\
\cline{1-5}
$0\%$ & $15.0\ \left(\pm 6.6\right)$ & $16.8\ \left(\pm 7.1\right)$ & $18.4\ \left(\pm 7.6\right)$ & $19.9\ \left(\pm 8.1\right)$ \\
$10\%$ & $13.2\ \left(\pm 5.9\right)$ & $15.5\ \left(\pm 6.7\right)$ & $16.8\ \left(\pm 7.2\right)$ & $19.4\ \left(\pm 8.0\right)$ \\
$20\%$ & $11.8\ \left(\pm 5.5\right)$ & $14.2\ \left(\pm 6.3\right)$ & $15.9\ \left(\pm 7.1\right)$ & $18.1\ \left(\pm 7.8\right)$ \\
$30\%$ & $11.2\ \left(\pm 5.4\right)$ & $13.5\ \left(\pm 6.3\right)$ & $15.5\ \left(\pm 7.1\right)$ & $17.3\ \left(\pm 7.8\right)$ \\
$40\%$ & $11.1\ \left(\pm 5.4\right)$ & $13.4\ \left(\pm 6.2\right)$ & $15.4\ \left(\pm 7.0\right)$ & $17.2\ \left(\pm 7.8\right)$ \\
\cline{1-5}
\end{tabular}
\end{table}

\begin{figure*}[tb]
    \centering
    \includegraphics[width=\textwidth]{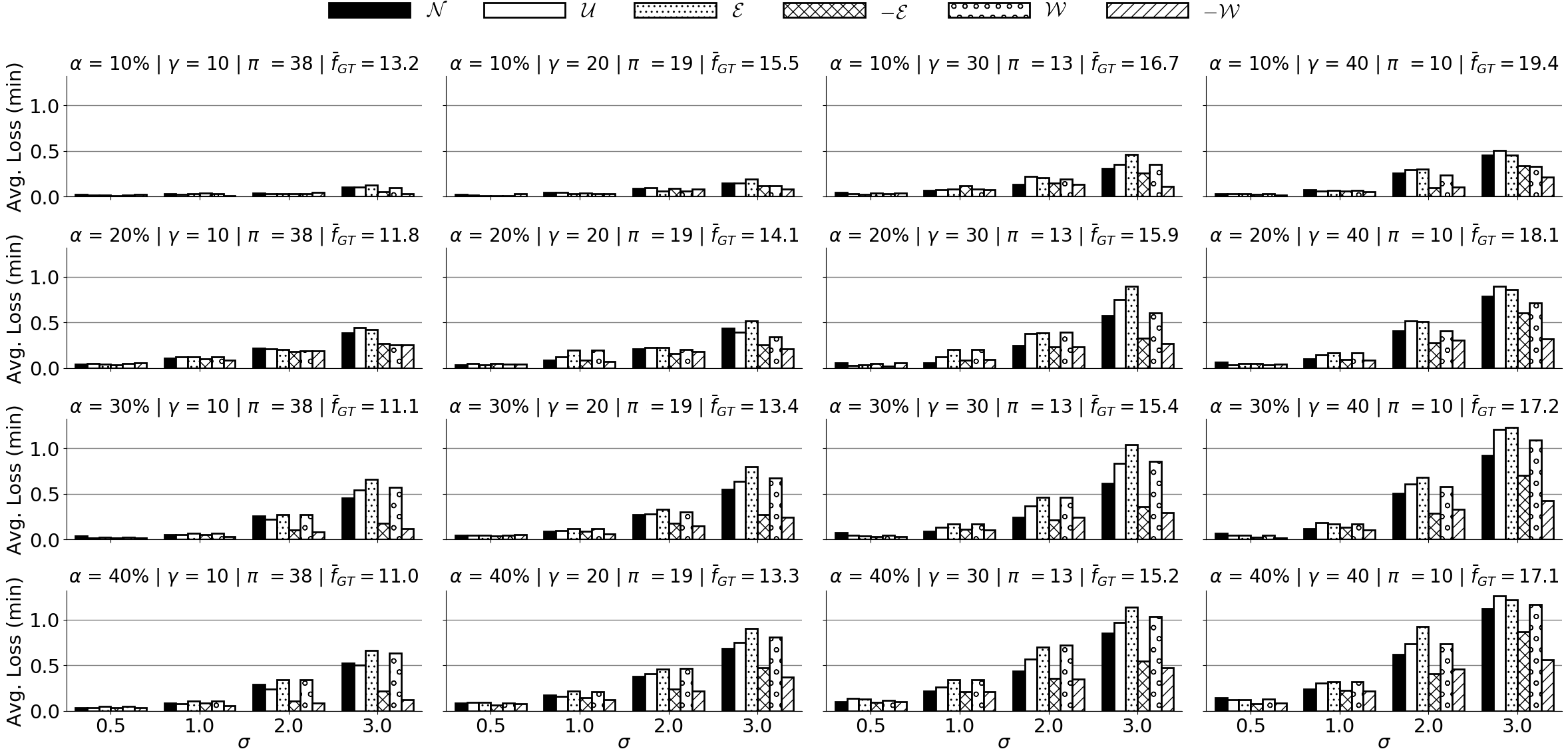}
    \caption{Average minutes lost with noisy vs. perfect predictions ($\bar{f}_{GT}$ in plot titles).
    Lower is better.
    Numeric results are in \tabapxref{tab:absloss}.} \label{fig:absloss}
\end{figure*}

We now proceed to measure how much time per passenger is theoretically lost when optimizing with noisy vs. perfect predictions, as:
\begin{equation}
    \absloss \coloneqq \frac{1}{N} \sum_{i=1}^{N} \frac{\opteval - \optevalGT}{P_i}
    \,.
    \label{eq:absloss}
\end{equation}
\Cref{fig:absloss} illustrates $\absloss$ using a grid of plots, where rows are ordered by percentage of dynamic buses ($\dperc$) and columns are ordered by bus capacity ($\capacity$).
We see that $\absloss$ increases when either $\SD$ or $\capacity$ increases, as expected.
We also see that for any fixed $\SD > 0$, a partial order on $\dstr$ emerges as $\capacity$ and $\dperc$ increase, so that $\negweib$ and $\negexp$ are significantly better than all other $\dstr$, and $\negweib < \negexp < \normal$.
These properties hold similarly for RMSNE but not MAPE, as seen earlier in \Cref{fig:predaftr}.
{
By properties of RMSNE, this suggests that as more dynamically routed buses are involved, the quality of routing becomes more sensitive to the presence of even a few large prediction errors -- i.e., consistent predictive accuracy is indeed essential.
}

Interestingly, \autoref{fig:absloss} also shows that $\absloss$ generally increases when $\dperc$ increases, i.e., when more dynamic buses are used for the same fleet size.
This holds also when normalizing $\tloss$ by the theoretical trip time with perfect predictions, namely:
\begin{equation}
    \relloss \coloneqq \frac{1}{N} \sum_{i=1}^{N} \frac{\opteval - \optevalGT}{\optevalGT}
    \,,
    \label{eq:relloss}
\end{equation}
as detailed in \tabapxref{tab:relloss}.
Still, $\tloss$ and $\tloss^{\text{rel}}$ are purely theoretical measurements, because observations cannot realistically be used before they manifest, and the predictive mean rarely captures them perfectly.

\begin{figure*}[tb]
    \includegraphics[width=\textwidth]{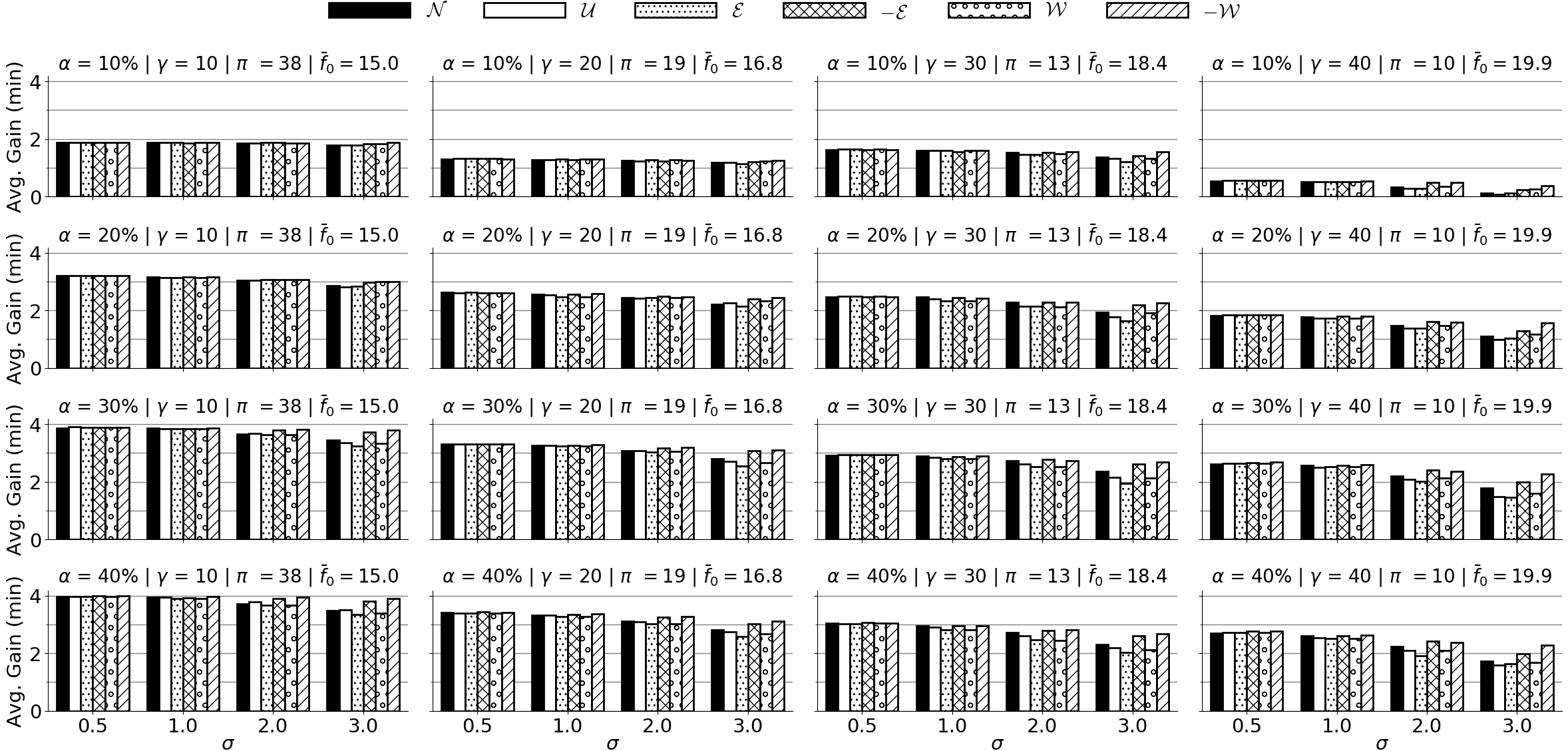}
    \caption{Average minutes gained with dynamic vs. completely static routing ($\bar{f}_0$ in plot titles).
    Higher is better.
    Numeric results are in \tabapxref{tab:absgain}.} \label{fig:absgain}
\end{figure*}

Next, we measure the time saved per passenger when using dynamic buses ($\dperc > 0$) vs. a completely statically routed fleet ($\dperc = 0$), namely:
\begin{equation}
    \absgain \coloneqq \frac{1}{N} \sum_{i=1}^{N} \frac{\optevalstatic - \opteval}{P_i}
    \,,
    \label{eq:absgain}
\end{equation}
as illustrated in \Cref{fig:absgain}.
We see that $\tgain$ generally improves as capacity and noise SD decrease and the percent of dynamic buses increases.
The only exception is $\tgain\left(\dstr, \SD, 20, 10\%\right) > \tgain\left(\dstr, \SD, 30, 20\%\right)$, for any $\dstr$ and $\SD$.
This is explained by \Cref{tab:ds}: for $\dperc = 10\%$, both $\capacity = 20, 30$ have the same no. dynamic buses, yet there are more statically routed buses for $\capacity = 20$ than for $\capacity = 30$.

We also see in \autoref{fig:absgain} that as $\capacity$ and $\dperc$ increase for any $\SD$, a partial order on $\dstr$ again emerges, with the same properties detected above for $\tloss$ and RMSNE.
{
This further supports our conclusion, that the consistency of predictive accuracy becomes more essential as the fleet becomes increasingly dynamic.
}
The best is $\tgain\left(\dstr, \SD, 10, 40\%\right) = \SI{4}{\min}$, for $\SD \leq 0.5$ and any $\dstr$.
The lowest is $\tgain\left(\dstr, 3.0, 40, 10\%\right) = \SI{0.1}{\min}$, for $\dstr \in \{\normal, \uni, \expon\}$.

We further convert $\tgain$ to an economic measure per the Danish Value of Travel Time Savings (VTTS), which has recently been estimated at $\vtts = \SI{13.43}{\EUR\per\hour}$ \citep{rich2019value}.
For each $\dperc$ and $\capacity$, we thus take the smallest $\absgain$ over all $\dstr\,,\SD$ and multiply it by both $\vtts$ (in $\si{\EUR\per\minute}$) and average no. trips per year in the studied OD pairs.
\Cref{tab:mineuro} provides the results, rounded to $\SI{1000}{\EUR\per\year}$, where the best yearly gains are at least $\SI{809000}{\EUR}$ (for $\dperc = 40\%, \capacity = 10$).

\begin{table}[tb]
    \centering
    \caption{Minimum yearly economic gains (\euro). Higher is better.}
    \label{tab:mineuro}  
\begin{tabular}{|c|c|c|c|c|}
\cline{1-5}
    $\dperc$ & $\capacity=10$ & $\capacity=20$ & $\capacity=30$ & $\capacity=40$ \\
\cline{1-5}
$10\%$ & $\num{429000}$ & $\num{274000}$ & $\num{293000}$ & $\hspace{4pt}\num{17000}$ \\
$20\%$ & $\num{677000}$ & $\num{515000}$ & $\num{392000}$ & $\num{238000}$ \\
$30\%$ & $\num{783000}$ & $\num{615000}$ & $\num{469000}$ & $\num{354000}$ \\
$40\%$ & $\num{809000}$ & $\num{623000}$ & $\num{488000}$ & $\num{382000}$ \\
\cline{1-5}
\end{tabular}
\end{table}

\begin{figure}[tb]
    \centering
    \includegraphics[width=0.5\linewidth]{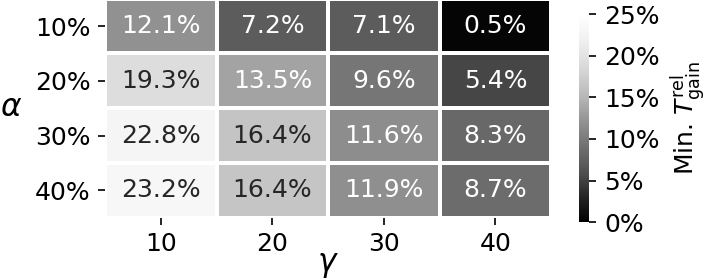}
    \caption{Minimum $\relgain$ over all $\dstr\,,\SD$. Higher is better.}
    \label{fig:minrelgain}
\end{figure}

Lastly, we measure the average relative gain when using a dynamic vs. completely static fleet, as:
\begin{equation}
    \relgain \coloneqq \frac{1}{N} \sum_{i=1}^{N} \frac{\optevalstatic - \opteval}{\optevalstatic}
    \,.
    \label{eq:relgain}
\end{equation}
This is detailed in Table \ref{tab:relgain}, where in particular, the best values of $\tgain^{\text{rel}}$ are close to $27.5\%$ while the worst are close to $5\%$.
As a concise reference for choosing operational parameters, \Cref{fig:minrelgain} summarizes $\tgain^{\text{rel}}$ in terms of its minimum over all noise distributions.
For instance, if operational parameters are chosen conservatively, so that each bus is large ($\capacity = 40$) and only $\dperc = 20\%$ of buses are dynamic, then at least $5\%$ relative gain is achieved.
However, if the fleet consists of vehicles with low capacity ($\capacity = 10$), many of which are dynamic ($\dperc = 40\%$), then at least $23\%$ relative gain is achieved.

\section{Discussion} \label{sec:discuss}

The main goal of this work is to quantify the effects of demand prediction accuracy on the performance of demand-responsive Public Transport (PT).
For this, we have used hourly observations of PT trips as a proxy for transport demand,
and conducted simulation experiments in two steps.
First, we have simulated the output of demand prediction models by perturbing the observations per various distributions, which cover a wide range of statistic properties.
Based on these noisy predictions, we have then used a mixed-integer linear program (MILP) with commonly used objective and constraints, to simulate and optimize PT fleets with varying no. statically and dynamically routed buses.

We have obtained that the differences in noise distributions do not account for much variability in trip time per passenger.
However, the noise distributions differ noticeably in two other measures of time per passenger: 1) time theoretically lost with noisy vs. perfect predictions, 2) time gained with dynamic vs. completely static routing.
The worst loss per passenger is $\SI{1.3}{\min}$, while the best time gain per passenger is $\SI{4}{\min}$, which is more than $27\%$ in relative terms.
In economic terms of Value of Time Savings (VTTS), the best gains in this case study are at least  $\SI{809000}{\EUR\per\year}$.

Also in terms of time gains and losses, we have obtained that the noise distributions rank more similarly to Rooted Squared Mean Normalized Error (RMSNE) than Mean Absolute Percentage Error (MAPE) of predictions.
As seen in \eqref{eq:mape} and \eqref{eq:rmsne}, RMSNE is dominated by exceptionally large prediction errors due to squaring, unlike MAPE.
It thus appears that exceptionally large prediction errors, even if few, can strongly influence the performance of dynamic PT optimization.

Finally, we have seen that when the common normality assumption is violated, optimization performance can not only worsen but also improve.
E.g., compared to the average gains and losses of the Gaussian distribution ($\normal$), the Uniform ($\uni$) and Weibull ($\weib$) are mostly worse, whereas the Negated Weibull ($\negweib$) and Negated Exponetial ($\negexp$) are mostly better.
In conjunction with \Cref{tab:skewkurt}, we find that this corresponds well to skew (rather than kurtosis), as $\uni$ and $\weib$ have positive skew, while $\negweib$ and $\negexp$ have negative skew.

\section{Conclusion} \label{sec:conclude}

In conclusion, the key findings of this work are as follows.
\begin{enumerate}
    \item {As a PT fleet becomes increasingly dynamic, the consistency of predictive accuracy becomes more essential for an effective service, i.e., to decrease temporal and economical losses and increase corresponding gains.}
    \item However, these gains still generally increase when using more dynamically routed buses (i.e., higher $\dperc$) with less capacity (i.e., lower $\capacity$), regardless of noise distribution.
    \item The minimum relative gain is $5\%$ when conservatively choosing $\dperc=20\%$ and $\capacity=40$ vs. $23\%$ when more liberally choosing $\dperc=10\%$ and $\capacity=10$.
    \autoref{fig:minrelgain} gives a fuller reference for choosing these operational parameters.
    \item 
    {
    Violations of the common normality assumption can in fact result in more reliable predictions, e.g., as in the case of error distributions with a negative skew.
    This encourages further research to identify conditions where predictive errors can actually improve predictive quality.
    }    
\end{enumerate}

For future work, we plan to enhance the current study in several respects, as follows.
We plan to also study cases where the noise distribution incorporates correlations between the OD's, by constructing their joint distribution \citep{peled2019online}.
We also plan to extend to the more general case of a full-scale PT fleet that serves more stations, and study how the predictive errors can be mitigated via robust optimization with chance constraints \citep{wang2015sustainable}.
We can also incorporate travelers' route choice behaviour by extending to a bi-modal optimization formulation {\citep{jiang2021reliability,jiang2021incorporating}}.
We further plan to incorporate truncated noise distributions \citep{greene2006censored}, to account for the inherently limited observability of PT usage \citep{gammelli2020estimating}, {as well as investigate conditions where predictive errors improve predictive quality}.
Given more computational resources, we also plan to measure the effects of higher percentages of dynamically routed buses on trip time minimization.
We can also consider a stochastic optimization formulation with recourse actions, to explicitly account for consequences of decisions made before the true travel demand is realized.

\section*{Declaration of competing interest}

The authors declare that they have no known competing financial interests or personal relationships that could have appeared to influence the work reported in this paper. 

\section*{Acknowledgements}

Data for this research was obtained by kind permission of the Danish Transport, Construction and Housing Authority, Movia Transit Agency, the Danish National Rail Company and the Danish Metro Company.

\bibliographystyle{myapa}
\bibliography{jourfull,newref}

\section*{Author biographies}

\minibio{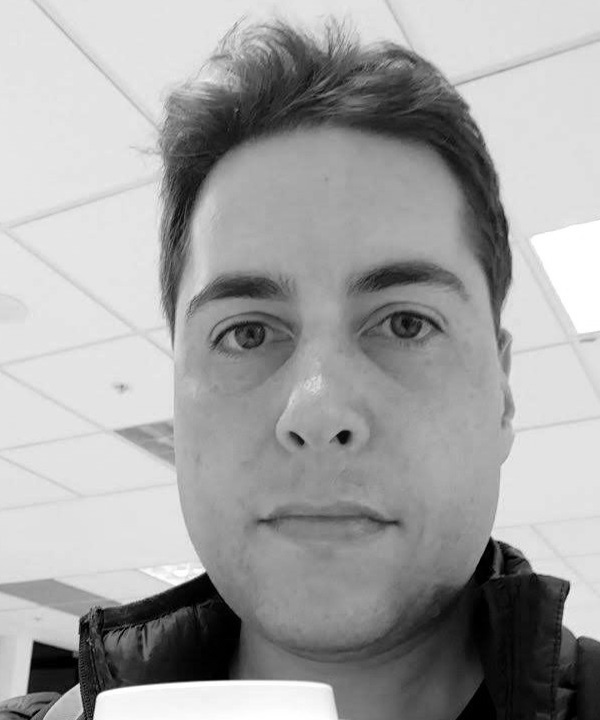}{Inon Peled}{is a Postdoctoral Researcher in the Machine Learning for Smart Mobility research group in the Technical University of Denmark (DTU), where he received his PhD in 2021.
His main research interests lie in predictive modeling for abnormal conditions in the transport domain.
}
\\
\minibio{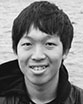}{Kelvin Lee}{is a PhD candidate with the Graduate College at Nanyang Technological University (NTU), Singapore.
He has received his BEng degree in electrical and electronic engineering from NTU.
His main research interests lie in predictive optimization methods with applications in transportation.
}
\\
\minibio{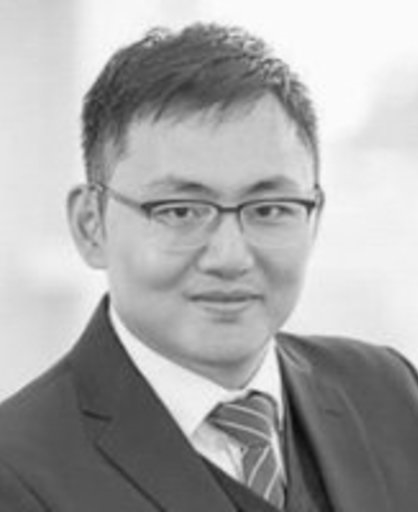}{Yu Jiang}{received his PhD in civil engineering from the University of Hong Kong (HKU) in 2014.
He is currently an Assoc. Professor in the Technical University of Denmark (DTU), specializing in modeling and optimizing public transport system.
}
\\
\minibio{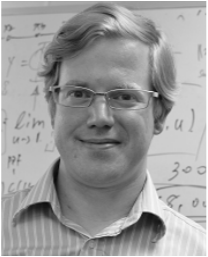}{Justin Dauwels}{received his PhD in electrical engineering from ETH, Zurich, in 2005.
He is an Assoc. Professor at the TU Delft.
His main research interests lie in probabilistic modeling, intelligent transportation systems and human behavior analysis.
He is also co-founder of several AI related spin-off companies.}
\\
\minibio{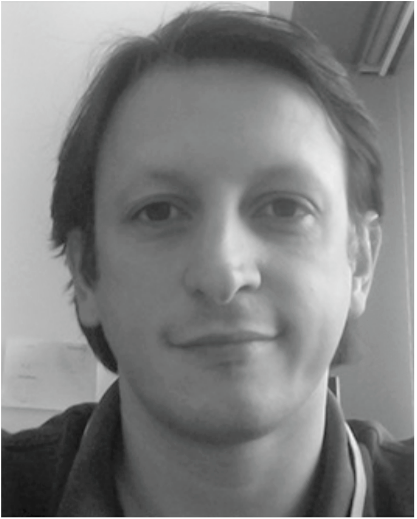}{Francisco C{\^a}mara Pereira}{received his PhD in computer science from the University of Coimbra, Portugal, in 2005.
He is a Full Professor and leader of the Machine Learning for Smart Mobility group in DTU. 
His main research interests lie in applying ML for predicting and improving transport systems.
}

\appendix
\section*{Appendix: numeric results} \label{sec:apxtables}

\vspace{70pt}

\setlength{\tabcolsep}{3.2pt}
\begin{table}[H]
\centering
\rotatebox{90}{%
  \begin{minipage}{0.65\textheight}
  \centering
  \caption{$\absloss$ in \si{\min}, lower is better.}\label{tab:absloss}
    \scriptsize
    \begin{tabular}{|cc|cccccc|cccccc|cccccc|cccccc|}
\cline{1-26}
     & & \multicolumn{6}{c|}{$\capacity=10$} & \multicolumn{6}{c|}{$\capacity=20$} & \multicolumn{6}{c|}{$\capacity=30$} & \multicolumn{6}{c|}{$\capacity=40$} \\
    $\dperc$ & $\SD$ & $\normal$ & $\uni$ & $\expon$ & $\negexp$ & $\weib$ & $\negweib$ & $\normal$ & $\uni$ & $\expon$ & $\negexp$ & $\weib$ & $\negweib$ & $\normal$ & $\uni$ & $\expon$ & $\negexp$ & $\weib$ & $\negweib$ & $\normal$ & $\uni$ & $\expon$ & $\negexp$ & $\weib$ & $\negweib$ \\
\cline{1-26}
\multirow{4}{*}{$10\%$} &  0.5 &     0.0 &      0.0 &    0.0 &     0.0 &      0.0 &      0.0 &     0.0 &      0.0 &    0.0 &     0.0 &      0.0 &      0.0 &     0.0 &      0.0 &    0.0 &     0.0 &      0.0 &      0.0 &     0.0 &      0.0 &    0.0 &     0.0 &      0.0 &      0.0 \\
    & 1.0 &     0.0 &      0.0 &    0.0 &     0.0 &      0.0 &      0.0 &     0.0 &      0.0 &    0.0 &     0.0 &      0.0 &      0.0 &     0.1 &      0.1 &    0.1 &     0.1 &      0.1 &      0.1 &     0.1 &      0.1 &    0.1 &     0.1 &      0.1 &      0.1 \\
    & 2.0 &     0.0 &      0.0 &    0.0 &     0.0 &      0.0 &      0.0 &     0.1 &      0.1 &    0.1 &     0.1 &      0.1 &      0.1 &     0.1 &      0.2 &    0.2 &     0.1 &      0.2 &      0.1 &     0.3 &      0.3 &    0.3 &     0.1 &      0.2 &      0.1 \\
    & 3.0 &     0.1 &      0.1 &    0.1 &     0.1 &      0.1 &      0.0 &     0.1 &      0.1 &    0.2 &     0.1 &      0.1 &      0.1 &     0.3 &      0.4 &    0.5 &     0.3 &      0.3 &      0.1 &     0.5 &      0.5 &    0.5 &     0.3 &      0.3 &      0.2 \\
\cline{1-26}
\multirow{4}{*}{$20\%$} & 0.5 &     0.0 &      0.0 &    0.0 &     0.0 &      0.0 &      0.1 &     0.0 &      0.0 &    0.0 &     0.0 &      0.0 &      0.0 &     0.1 &      0.0 &    0.0 &     0.1 &      0.0 &      0.1 &     0.1 &      0.0 &    0.0 &     0.0 &      0.0 &      0.0 \\
    & 1.0 &     0.1 &      0.1 &    0.1 &     0.1 &      0.1 &      0.1 &     0.1 &      0.1 &    0.2 &     0.1 &      0.2 &      0.1 &     0.1 &      0.1 &    0.2 &     0.1 &      0.2 &      0.1 &     0.1 &      0.1 &    0.2 &     0.1 &      0.2 &      0.1 \\
    & 2.0 &     0.2 &      0.2 &    0.2 &     0.2 &      0.2 &      0.2 &     0.2 &      0.2 &    0.2 &     0.2 &      0.2 &      0.2 &     0.2 &      0.4 &    0.4 &     0.2 &      0.4 &      0.2 &     0.4 &      0.5 &    0.5 &     0.3 &      0.4 &      0.3 \\
    & 3.0 &     0.4 &      0.4 &    0.4 &     0.3 &      0.3 &      0.3 &     0.4 &      0.4 &    0.5 &     0.3 &      0.3 &      0.2 &     0.6 &      0.7 &    0.9 &     0.3 &      0.6 &      0.3 &     0.8 &      0.9 &    0.9 &     0.6 &      0.7 &      0.3 \\
\cline{1-26}
\multirow{4}{*}{$30\%$} & 0.5 &     0.0 &      0.0 &    0.0 &     0.0 &      0.0 &      0.0 &     0.0 &      0.0 &    0.0 &     0.0 &      0.0 &      0.1 &     0.1 &      0.0 &    0.0 &     0.0 &      0.0 &      0.0 &     0.1 &      0.0 &    0.0 &     0.0 &      0.0 &      0.0 \\
    & 1.0 &     0.1 &      0.1 &    0.1 &     0.1 &      0.1 &      0.0 &     0.1 &      0.1 &    0.1 &     0.1 &      0.1 &      0.1 &     0.1 &      0.1 &    0.2 &     0.1 &      0.2 &      0.1 &     0.1 &      0.2 &    0.2 &     0.1 &      0.2 &      0.1 \\
    & 2.0 &     0.3 &      0.2 &    0.3 &     0.1 &      0.3 &      0.1 &     0.3 &      0.3 &    0.3 &     0.2 &      0.3 &      0.1 &     0.2 &      0.4 &    0.5 &     0.2 &      0.5 &      0.2 &     0.5 &      0.6 &    0.7 &     0.3 &      0.6 &      0.3 \\
    & 3.0 &     0.5 &      0.5 &    0.7 &     0.2 &      0.6 &      0.1 &     0.5 &      0.6 &    0.8 &     0.3 &      0.7 &      0.2 &     0.6 &      0.8 &    1.0 &     0.4 &      0.9 &      0.3 &     0.9 &      1.2 &    1.2 &     0.7 &      1.1 &      0.4 \\
\cline{1-26}
\multirow{4}{*}{$40\%$} & 0.5 &     0.0 &      0.0 &    0.0 &     0.0 &      0.0 &      0.0 &     0.1 &      0.1 &    0.1 &     0.1 &      0.1 &      0.1 &     0.1 &      0.1 &    0.1 &     0.1 &      0.1 &      0.1 &     0.1 &      0.1 &    0.1 &     0.1 &      0.1 &      0.1 \\
    & 1.0 &     0.1 &      0.1 &    0.1 &     0.1 &      0.1 &      0.1 &     0.2 &      0.2 &    0.2 &     0.1 &      0.2 &      0.1 &     0.2 &      0.3 &    0.3 &     0.2 &      0.3 &      0.2 &     0.2 &      0.3 &    0.3 &     0.2 &      0.3 &      0.2 \\
    & 2.0 &     0.3 &      0.2 &    0.3 &     0.1 &      0.3 &      0.1 &     0.4 &      0.4 &    0.5 &     0.2 &      0.5 &      0.2 &     0.4 &      0.6 &    0.7 &     0.4 &      0.7 &      0.3 &     0.6 &      0.7 &    0.9 &     0.4 &      0.7 &      0.5 \\
    & 3.0 &     0.5 &      0.5 &    0.7 &     0.2 &      0.6 &      0.1 &     0.7 &      0.7 &    0.9 &     0.5 &      0.8 &      0.4 &     0.9 &      1.0 &    1.1 &     0.5 &      1.0 &      0.5 &     1.1 &      1.3 &    1.2 &     0.9 &      1.2 &      0.6 \\
\cline{1-26}
\end{tabular}
  \end{minipage}%
}
\end{table}

\begin{table}[tb]
\centering
\rotatebox{90}{%
  \begin{minipage}{0.89\textheight}
  \centering
  \caption{$\relloss$ in \%, lower is better.}\label{tab:relloss}
    \scriptsize
\begin{tabular}{|cc|cccccc|cccccc|cccccc|cccccc|}
\cline{1-26}
    & & \multicolumn{6}{c|}{$\gamma=10$} & \multicolumn{6}{c|}{$\gamma=20$} & \multicolumn{6}{c|}{$\gamma=30$} & \multicolumn{6}{c|}{$\gamma=40$} \\
    $\alpha$ & $\sigma$ & $\normal$ & $\uni$ & $\expon$ & $\negexp$ & $\weib$ & $\negweib$ & $\normal$ & $\uni$ & $\expon$ & $\negexp$ & $\weib$ & $\negweib$ & $\normal$ & $\uni$ & $\expon$ & $\negexp$ & $\weib$ & $\negweib$ & $\normal$ & $\uni$ & $\expon$ & $\negexp$ & $\weib$ & $\negweib$ \\
\cline{1-26}
\multirow{4}{*}{$10\%$} & 0.5 &    0.1 &     0.1 &   0.1 &    0.1 &     0.1 &     0.1 &    0.1 &     0.1 &   0.1 &    0.1 &     0.1 &     0.1 &    0.2 &     0.2 &   0.1 &    0.2 &     0.2 &     0.2 &    0.1 &     0.2 &   0.2 &    0.1 &     0.2 &     0.1 \\
    & 1.0 &    0.1 &     0.2 &   0.2 &    0.3 &     0.2 &     0.1 &    0.3 &     0.3 &   0.2 &    0.2 &     0.2 &     0.3 &    0.4 &     0.5 &   0.5 &    0.7 &     0.5 &     0.4 &    0.3 &     0.3 &   0.4 &    0.3 &     0.4 &     0.2 \\
    & 2.0 &    0.3 &     0.3 &   0.2 &    0.3 &     0.3 &     0.3 &    0.5 &     0.7 &   0.4 &    0.5 &     0.4 &     0.5 &    0.7 &     1.4 &   1.2 &    0.9 &     1.1 &     0.7 &    1.4 &     1.5 &   1.6 &    0.5 &     1.3 &     0.5 \\
    & 3.0 &    0.7 &     0.6 &   0.8 &    0.5 &     0.7 &     0.2 &    0.8 &     0.7 &   1.1 &    0.9 &     0.7 &     0.5 &    1.8 &     1.9 &   2.6 &    1.6 &     2.1 &     0.6 &    2.5 &     2.8 &   2.4 &    2.0 &     1.8 &     1.1 \\
\cline{1-26}
\multirow{4}{*}{$20\%$} & 0.5 &    0.3 &     0.3 &   0.3 &    0.3 &     0.3 &     0.4 &    0.3 &     0.3 &   0.2 &    0.3 &     0.3 &     0.2 &    0.2 &     0.1 &   0.1 &    0.2 &     0.1 &     0.2 &    0.3 &     0.2 &   0.2 &    0.2 &     0.2 &     0.2 \\
    & 1.0 &    1.0 &     1.1 &   1.1 &    0.9 &     1.1 &     0.8 &    0.6 &     0.9 &   1.1 &    0.5 &     1.1 &     0.5 &    0.3 &     0.7 &   1.0 &    0.4 &     1.0 &     0.4 &    0.5 &     0.8 &   0.8 &    0.4 &     0.8 &     0.4 \\
    & 2.0 &    2.0 &     2.1 &   2.0 &    1.8 &     1.8 &     1.8 &    1.4 &     1.9 &   1.7 &    1.2 &     1.5 &     1.2 &    1.5 &     2.6 &   2.3 &    1.4 &     2.3 &     1.4 &    2.3 &     3.0 &   2.9 &    1.4 &     2.4 &     1.5 \\
    & 3.0 &    3.3 &     3.7 &   3.7 &    2.6 &     2.2 &     2.4 &    3.0 &     2.7 &   3.4 &    1.9 &     2.2 &     1.6 &    3.6 &     4.6 &   5.5 &    2.0 &     3.8 &     1.7 &    4.6 &     5.0 &   5.1 &    3.4 &     4.0 &     1.8 \\
\cline{1-26}
\multirow{4}{*}{$30\%$} & 0.5 &    0.2 &     0.1 &   0.2 &    0.1 &     0.1 &     0.1 &    0.2 &     0.3 &   0.3 &    0.2 &     0.3 &     0.3 &    0.3 &     0.2 &   0.2 &    0.1 &     0.2 &     0.1 &    0.2 &     0.2 &   0.2 &    0.1 &     0.2 &     0.1 \\
    & 1.0 &    0.4 &     0.5 &   0.6 &    0.4 &     0.6 &     0.2 &    0.6 &     0.8 &   0.7 &    0.6 &     0.7 &     0.4 &    0.5 &     0.8 &   0.8 &    0.5 &     0.8 &     0.5 &    0.6 &     1.0 &   0.8 &    0.5 &     0.8 &     0.5 \\
    & 2.0 &    2.3 &     2.3 &   2.5 &    1.0 &     2.4 &     0.7 &    1.9 &     2.5 &   2.5 &    1.3 &     2.2 &     1.0 &    1.5 &     2.8 &   3.0 &    1.3 &     2.9 &     1.4 &    3.0 &     3.7 &   3.9 &    1.6 &     3.3 &     1.6 \\
    & 3.0 &    4.2 &     4.8 &   5.5 &    1.8 &     4.9 &     1.1 &    3.8 &     4.3 &   5.5 &    2.2 &     4.8 &     1.9 &    4.1 &     5.0 &   6.5 &    2.4 &     5.4 &     2.0 &    5.9 &     7.1 &   7.1 &    4.1 &     6.2 &     2.7 \\
\cline{1-26}
\multirow{4}{*}{$40\%$} & 0.5 &    0.2 &     0.2 &   0.3 &    0.2 &     0.3 &     0.2 &    0.5 &     0.5 &   0.6 &    0.3 &     0.6 &     0.4 &    0.5 &     0.7 &   0.7 &    0.5 &     0.6 &     0.6 &    0.6 &     0.6 &   0.6 &    0.4 &     0.6 &     0.4 \\
    & 1.0 &    0.6 &     0.7 &   0.9 &    0.5 &     0.9 &     0.4 &    1.1 &     1.1 &   1.4 &    0.9 &     1.4 &     0.8 &    1.3 &     1.5 &   1.8 &    1.1 &     1.8 &     1.1 &    1.2 &     1.6 &   1.7 &    1.1 &     1.7 &     1.1 \\
    & 2.0 &    2.5 &     2.2 &   3.2 &    0.9 &     3.2 &     0.7 &    2.6 &     3.2 &   3.4 &    1.7 &     3.4 &     1.4 &    2.6 &     3.8 &   4.3 &    2.0 &     4.5 &     2.0 &    3.6 &     4.4 &   5.4 &    2.3 &     4.3 &     2.3 \\
    & 3.0 &    4.8 &     4.3 &   6.0 &    2.1 &     5.8 &     1.0 &    4.9 &     5.4 &   6.5 &    3.6 &     6.0 &     2.8 &    5.6 &     6.0 &   7.2 &    3.5 &     6.7 &     2.9 &    7.1 &     7.5 &   7.1 &    5.0 &     6.8 &     3.4 \\
\cline{1-26}
\end{tabular}
  \end{minipage}%
}
\end{table}

\begin{table}[tb]
\centering
\rotatebox{90}{%
  \begin{minipage}{0.89\textheight}
  \centering
  \caption{$\absgain$ in \si{\min}, higher is better.}\label{tab:absgain}
    \scriptsize
\begin{tabular}{|cc|cccccc|cccccc|cccccc|cccccc|}
\cline{1-26}
     & & \multicolumn{6}{c|}{$\capacity=10$} & \multicolumn{6}{c|}{$\capacity=20$} & \multicolumn{6}{c|}{$\capacity=30$} & \multicolumn{6}{c|}{$\capacity=40$} \\
    $\dperc$ & $\SD$ & $\normal$ & $\uni$ & $\expon$ & $\negexp$ & $\weib$ & $\negweib$ & $\normal$ & $\uni$ & $\expon$ & $\negexp$ & $\weib$ & $\negweib$ & $\normal$ & $\uni$ & $\expon$ & $\negexp$ & $\weib$ & $\negweib$ & $\normal$ & $\uni$ & $\expon$ & $\negexp$ & $\weib$ & $\negweib$ \\
\cline{1-26}
\multirow{5}{*}{$10\%$} & 0.0 &     1.9 &      1.9 &    1.9 &     1.9 &      1.9 &      1.9 &     1.3 &      1.3 &    1.3 &     1.3 &      1.3 &      1.3 &     1.7 &      1.7 &    1.7 &     1.7 &      1.7 &      1.7 &     0.6 &      0.6 &    0.6 &     0.6 &      0.6 &      0.6 \\
    & 0.5 &     1.9 &      1.9 &    1.9 &     1.9 &      1.9 &      1.9 &     1.3 &      1.3 &    1.3 &     1.3 &      1.3 &      1.3 &     1.6 &      1.6 &    1.7 &     1.6 &      1.6 &      1.6 &     0.5 &      0.5 &    0.5 &     0.6 &      0.6 &      0.6 \\
    & 1.0 &     1.9 &      1.9 &    1.9 &     1.9 &      1.9 &      1.9 &     1.3 &      1.3 &    1.3 &     1.3 &      1.3 &      1.3 &     1.6 &      1.6 &    1.6 &     1.6 &      1.6 &      1.6 &     0.5 &      0.5 &    0.5 &     0.5 &      0.5 &      0.5 \\
    & 2.0 &     1.9 &      1.9 &    1.9 &     1.9 &      1.9 &      1.8 &     1.2 &      1.2 &    1.3 &     1.2 &      1.3 &      1.2 &     1.5 &      1.5 &    1.5 &     1.5 &      1.5 &      1.5 &     0.3 &      0.3 &    0.3 &     0.5 &      0.3 &      0.5 \\
    & 3.0 &     1.8 &      1.8 &    1.8 &     1.8 &      1.8 &      1.9 &     1.2 &      1.2 &    1.1 &     1.2 &      1.2 &      1.2 &     1.4 &      1.3 &    1.2 &     1.4 &      1.3 &      1.6 &     0.1 &      0.1 &    0.1 &     0.2 &      0.2 &      0.4 \\
\cline{1-26}
\multirow{5}{*}{$20\%$} & 0.0 &     3.3 &      3.3 &    3.3 &     3.3 &      3.3 &      3.3 &     2.7 &      2.7 &    2.7 &     2.7 &      2.7 &      2.7 &     2.5 &      2.5 &    2.5 &     2.5 &      2.5 &      2.5 &     1.9 &      1.9 &    1.9 &     1.9 &      1.9 &      1.9 \\
    & 0.5 &     3.2 &      3.2 &    3.2 &     3.2 &      3.2 &      3.2 &     2.6 &      2.6 &    2.6 &     2.6 &      2.6 &      2.6 &     2.5 &      2.5 &    2.5 &     2.5 &      2.5 &      2.5 &     1.8 &      1.8 &    1.8 &     1.8 &      1.8 &      1.8 \\
    & 1.0 &     3.2 &      3.1 &    3.1 &     3.2 &      3.1 &      3.2 &     2.6 &      2.5 &    2.5 &     2.6 &      2.5 &      2.6 &     2.5 &      2.4 &    2.3 &     2.4 &      2.3 &      2.4 &     1.8 &      1.7 &    1.7 &     1.8 &      1.7 &      1.8 \\
    & 2.0 &     3.0 &      3.0 &    3.1 &     3.1 &      3.1 &      3.1 &     2.4 &      2.4 &    2.4 &     2.5 &      2.5 &      2.5 &     2.3 &      2.1 &    2.1 &     2.3 &      2.1 &      2.3 &     1.5 &      1.4 &    1.4 &     1.6 &      1.5 &      1.6 \\
    & 3.0 &     2.9 &      2.8 &    2.8 &     3.0 &      3.0 &      3.0 &     2.2 &      2.3 &    2.1 &     2.4 &      2.3 &      2.4 &     1.9 &      1.8 &    1.6 &     2.2 &      1.9 &      2.3 &     1.1 &      1.0 &    1.0 &     1.3 &      1.2 &      1.6 \\
\cline{1-26}
\multirow{5}{*}{$30\%$} & 0.0 &     3.9 &      3.9 &    3.9 &     3.9 &      3.9 &      3.9 &     3.4 &      3.4 &    3.4 &     3.4 &      3.4 &      3.4 &     3.0 &      3.0 &    3.0 &     3.0 &      3.0 &      3.0 &     2.7 &      2.7 &    2.7 &     2.7 &      2.7 &      2.7 \\
    & 0.5 &     3.9 &      3.9 &    3.9 &     3.9 &      3.9 &      3.9 &     3.3 &      3.3 &    3.3 &     3.3 &      3.3 &      3.3 &     2.9 &      2.9 &    2.9 &     3.0 &      2.9 &      3.0 &     2.6 &      2.6 &    2.7 &     2.7 &      2.7 &      2.7 \\
    & 1.0 &     3.9 &      3.9 &    3.8 &     3.9 &      3.8 &      3.9 &     3.3 &      3.3 &    3.2 &     3.3 &      3.2 &      3.3 &     2.9 &      2.9 &    2.8 &     2.9 &      2.8 &      2.9 &     2.6 &      2.5 &    2.5 &     2.6 &      2.5 &      2.6 \\
    & 2.0 &     3.7 &      3.7 &    3.6 &     3.8 &      3.6 &      3.8 &     3.1 &      3.1 &    3.0 &     3.2 &      3.1 &      3.2 &     2.7 &      2.6 &    2.5 &     2.8 &      2.5 &      2.7 &     2.2 &      2.1 &    2.0 &     2.4 &      2.1 &      2.4 \\
    & 3.0 &     3.5 &      3.4 &    3.2 &     3.7 &      3.3 &      3.8 &     2.8 &      2.7 &    2.6 &     3.1 &      2.7 &      3.1 &     2.4 &      2.2 &    1.9 &     2.6 &      2.1 &      2.7 &     1.8 &      1.5 &    1.5 &     2.0 &      1.6 &      2.3 \\
\cline{1-26}
\multirow{5}{*}{$40\%$} & 0.0 &     4.0 &      4.0 &    4.0 &     4.0 &      4.0 &      4.0 &     3.5 &      3.5 &    3.5 &     3.5 &      3.5 &      3.5 &     3.2 &      3.2 &    3.2 &     3.2 &      3.2 &      3.2 &     2.8 &      2.8 &    2.8 &     2.8 &      2.8 &      2.8 \\
    & 0.5 &     4.0 &      4.0 &    4.0 &     4.0 &      4.0 &      4.0 &     3.4 &      3.4 &    3.4 &     3.4 &      3.4 &      3.4 &     3.1 &      3.0 &    3.0 &     3.1 &      3.0 &      3.1 &     2.7 &      2.7 &    2.7 &     2.8 &      2.7 &      2.8 \\
    & 1.0 &     3.9 &      3.9 &    3.9 &     3.9 &      3.9 &      4.0 &     3.3 &      3.3 &    3.3 &     3.4 &      3.3 &      3.4 &     2.9 &      2.9 &    2.8 &     3.0 &      2.8 &      2.9 &     2.6 &      2.5 &    2.5 &     2.6 &      2.5 &      2.6 \\
    & 2.0 &     3.7 &      3.8 &    3.7 &     3.9 &      3.7 &      3.9 &     3.1 &      3.1 &    3.0 &     3.3 &      3.0 &      3.3 &     2.7 &      2.6 &    2.5 &     2.8 &      2.4 &      2.8 &     2.2 &      2.1 &    1.9 &     2.4 &      2.1 &      2.4 \\
    & 3.0 &     3.5 &      3.5 &    3.4 &     3.8 &      3.4 &      3.9 &     2.8 &      2.7 &    2.6 &     3.0 &      2.7 &      3.1 &     2.3 &      2.2 &    2.0 &     2.6 &      2.1 &      2.7 &     1.7 &      1.6 &    1.6 &     2.0 &      1.7 &      2.3 \\
\cline{1-26}
\end{tabular}
  \end{minipage}%
}
\end{table}

\begin{table}[tb]
\centering
\rotatebox{90}{%
  \begin{minipage}{0.89\textheight}
  \centering
  \caption{$\relgain$ in \%, higher is better.}\label{tab:relgain}
    \scriptsize
    \begin{tabular}{|cc|cccccc|cccccc|cccccc|cccccc|}
\cline{1-26}
     & & \multicolumn{6}{c|}{$\capacity=10$} & \multicolumn{6}{c|}{$\capacity=20$} & \multicolumn{6}{c|}{$\capacity=30$} & \multicolumn{6}{c|}{$\capacity=40$} \\
    $\dperc$ & $\SD$ & $\normal$ & $\uni$ & $\expon$ & $\negexp$ & $\weib$ & $\negweib$ & $\normal$ & $\uni$ & $\expon$ & $\negexp$ & $\weib$ & $\negweib$ & $\normal$ & $\uni$ & $\expon$ & $\negexp$ & $\weib$ & $\negweib$ & $\normal$ & $\uni$ & $\expon$ & $\negexp$ & $\weib$ & $\negweib$ \\
\cline{1-26}
\multirow{5}{*}{$10\%$} & 0.0 &    12.7 &     12.7 &   12.7 &    12.7 &     12.7 &     12.7 &    8.2 &     8.2 &   8.2 &    8.2 &     8.2 &     8.2 &    9.5 &     9.5 &   9.5 &    9.5 &     9.5 &     9.5 &    3.1 &     3.1 &   3.1 &    3.1 &     3.1 &     3.1 \\
    & 0.5 &    12.6 &     12.6 &   12.6 &    12.7 &     12.7 &     12.6 &    8.0 &     8.1 &   8.1 &    8.1 &     8.1 &     8.0 &    9.3 &     9.3 &   9.4 &    9.3 &     9.3 &     9.3 &    3.0 &     2.9 &   3.0 &    3.0 &     3.0 &     3.0 \\
    & 1.0 &    12.6 &     12.5 &   12.6 &    12.5 &     12.6 &     12.7 &    7.9 &     7.9 &   8.0 &    7.9 &     8.0 &     7.9 &    9.1 &     9.0 &   9.0 &    8.9 &     9.0 &     9.1 &    2.8 &     2.8 &   2.8 &    2.8 &     2.8 &     2.9 \\
    & 2.0 &    12.5 &     12.5 &   12.5 &    12.5 &     12.5 &     12.5 &    7.7 &     7.5 &   7.8 &    7.7 &     7.8 &     7.7 &    8.8 &     8.2 &   8.3 &    8.7 &     8.5 &     8.9 &    1.7 &     1.6 &   1.6 &    2.6 &     1.9 &     2.6 \\
    & 3.0 &    12.1 &     12.2 &   12.1 &    12.3 &     12.2 &     12.5 &    7.4 &     7.5 &   7.2 &    7.3 &     7.6 &     7.7 &    7.8 &     7.8 &   7.1 &    8.0 &     7.6 &     8.9 &    0.7 &     0.5 &   0.8 &    1.2 &     1.3 &     2.0 \\
\cline{1-26}
\multirow{5}{*}{$20\%$} & 0.0 &    22.2 &     22.2 &   22.2 &    22.2 &     22.2 &     22.2 &    16.3 &     16.3 &   16.3 &    16.3 &     16.3 &     16.3 &    14.3 &     14.3 &   14.3 &    14.3 &     14.3 &     14.3 &    10.0 &     10.0 &   10.0 &    10.0 &     10.0 &     10.0 \\
    & 0.5 &    22.0 &     21.9 &   22.0 &    22.0 &     21.9 &     21.9 &    16.1 &     16.1 &   16.1 &    16.1 &     16.1 &     16.1 &    14.1 &     14.2 &   14.2 &    14.1 &     14.2 &     14.1 &    9.8 &     9.9 &   9.8 &    9.8 &     9.9 &     9.9 \\
    & 1.0 &    21.5 &     21.3 &   21.4 &    21.5 &     21.4 &     21.6 &    15.8 &     15.6 &   15.4 &    15.9 &     15.4 &     15.9 &    14.1 &     13.7 &   13.5 &    14.0 &     13.5 &     13.9 &    9.6 &     9.3 &   9.3 &    9.7 &     9.3 &     9.7 \\
    & 2.0 &    20.7 &     20.6 &   20.7 &    20.9 &     20.8 &     20.9 &    15.2 &     14.8 &   14.9 &    15.3 &     15.1 &     15.3 &    13.1 &     12.1 &   12.3 &    13.1 &     12.4 &     13.2 &    8.0 &     7.3 &   7.5 &    8.8 &     7.9 &     8.7 \\
    & 3.0 &    19.7 &     19.3 &   19.4 &    20.2 &     20.5 &     20.3 &    13.9 &     14.1 &   13.5 &    14.7 &     14.4 &     15.0 &    11.3 &     10.4 &   9.6 &    12.6 &     11.0 &     12.9 &    5.9 &     5.5 &   5.4 &    7.0 &     6.4 &     8.4 \\
\cline{1-26}
\multirow{5}{*}{$30\%$} & 0.0 &    26.8 &     26.8 &   26.8 &    26.8 &     26.8 &     26.8 &    20.8 &     20.8 &   20.8 &    20.8 &     20.8 &     20.8 &    17.1 &     17.1 &   17.1 &    17.1 &     17.1 &     17.1 &    14.4 &     14.4 &   14.4 &    14.4 &     14.4 &     14.4 \\
    & 0.5 &    26.7 &     26.7 &   26.7 &    26.7 &     26.7 &     26.7 &    20.6 &     20.6 &   20.5 &    20.6 &     20.5 &     20.5 &    16.8 &     16.9 &   16.9 &    16.9 &     16.9 &     16.9 &    14.2 &     14.2 &   14.2 &    14.3 &     14.2 &     14.4 \\
    & 1.0 &    26.5 &     26.5 &   26.4 &    26.5 &     26.4 &     26.6 &    20.3 &     20.2 &   20.2 &    20.3 &     20.2 &     20.5 &    16.6 &     16.4 &   16.3 &    16.6 &     16.3 &     16.6 &    13.9 &     13.6 &   13.7 &    14.0 &     13.7 &     14.0 \\
    & 2.0 &    25.2 &     25.2 &   25.0 &    26.1 &     25.1 &     26.3 &    19.3 &     18.8 &   18.8 &    19.7 &     19.0 &     20.0 &    15.8 &     14.8 &   14.6 &    16.0 &     14.7 &     15.9 &    11.8 &     11.3 &   11.1 &    13.0 &     11.6 &     13.0 \\
    & 3.0 &    23.7 &     23.3 &   22.8 &    25.5 &     23.2 &     26.0 &    17.8 &     17.4 &   16.4 &    19.0 &     17.0 &     19.2 &    13.7 &     12.9 &   11.6 &    15.1 &     12.6 &     15.4 &    9.5 &     8.4 &   8.3 &    10.9 &     9.1 &     12.1 \\
\cline{1-26}
\multirow{5}{*}{$40\%$} & 0.0 &    27.5 &     27.5 &   27.5 &    27.5 &     27.5 &     27.5 &    21.5 &     21.5 &   21.5 &    21.5 &     21.5 &     21.5 &    17.9 &     17.9 &   17.9 &    17.9 &     17.9 &     17.9 &    15.1 &     15.1 &   15.1 &    15.1 &     15.1 &     15.1 \\
    & 0.5 &    27.3 &     27.3 &   27.3 &    27.4 &     27.3 &     27.4 &    21.1 &     21.1 &   21.0 &    21.2 &     21.0 &     21.2 &    17.4 &     17.3 &   17.3 &    17.4 &     17.3 &     17.4 &    14.5 &     14.6 &   14.6 &    14.7 &     14.5 &     14.7 \\
    & 1.0 &    27.1 &     27.1 &   26.9 &    27.1 &     26.9 &     27.2 &    20.6 &     20.7 &   20.4 &    20.8 &     20.4 &     20.9 &    16.8 &     16.7 &   16.4 &    17.0 &     16.4 &     16.9 &    14.0 &     13.7 &   13.6 &    14.2 &     13.6 &     14.1 \\
    & 2.0 &    25.7 &     26.0 &   25.3 &    26.9 &     25.2 &     27.0 &    19.5 &     19.0 &   18.8 &    20.2 &     18.9 &     20.4 &    15.7 &     14.7 &   14.3 &    16.2 &     14.2 &     16.2 &    12.0 &     11.4 &   10.6 &    13.1 &     11.5 &     13.1 \\
    & 3.0 &    24.0 &     24.4 &   23.2 &    26.0 &     23.3 &     26.8 &    17.7 &     17.3 &   16.4 &    18.7 &     16.8 &     19.4 &    13.3 &     12.9 &   11.9 &    15.0 &     12.3 &     15.5 &    9.2 &     8.7 &   9.0 &    10.8 &     9.3 &     12.2 \\
\cline{1-26}
\end{tabular}    
\end{minipage}%
}
\end{table}

\end{document}